\documentclass[letterpaper, 10 pt, conference]{ieeeconf} 

\IEEEoverridecommandlockouts                              

\usepackage{amsmath} 

\usepackage{amssymb}  
\usepackage{booktabs, tabularx, makecell}
\usepackage{amssymb}

\usepackage{bbm}
\usepackage[bb=boondox]{mathalfa}
\usepackage{graphicx} 
\usepackage{hyperref}
\usepackage{booktabs,arydshln}
\makeatletter
\def\adl@drawiv#1#2#3{%
        \hskip.5\tabcolsep
        \xleaders#3{#2.5\@tempdimb #1{1}#2.5\@tempdimb}%
                #2\z@ plus1fil minus1fil\relax
        \hskip.5\tabcolsep}
\newcommand{\cdashlinelr}[1]{%
  \noalign{\vskip\aboverulesep
           \global\let\@dashdrawstore\adl@draw
           \global\let\adl@draw\adl@drawiv}
  \cdashline{#1}
  \noalign{\global\let\adl@draw\@dashdrawstore
           \vskip\belowrulesep}}
\makeatother

\usepackage{threeparttable}
\usepackage{tabularx}
\usepackage{multirow}
\usepackage[caption=false, font=footnotesize]{subfig}
\usepackage{algorithmic}
\usepackage{algorithm}

\usepackage{color}
\usepackage{xcolor}
\usepackage{colortbl}
\usepackage[whole]{bxcjkjatype}
\definecolor{green}{rgb}{0.8,1.0,0.8}
\definecolor{red}{rgb}{1.0,0.8,0.8}

\usepackage{comment}
\usepackage{booktabs}
\usepackage{pifont}          

\newcolumntype{V}{!{\vrule width 0.6pt}}
\newcommand{\best}[1]{\cellcolor{black!10}\textbf{#1}}
\newcommand{\runup}[1]{\cellcolor{black!10}\underline{#1}} 

\definecolor{ctodo}{rgb}{1, 0., 0.}

\usepackage{xcolor}
\usepackage[normalem]{ulem}

\definecolor{DiffRed}{RGB}{180,0,0}

\newcommand{\Remove}[1]{}

\usepackage[%
backend=biber,
url=true,
block=space,
maxbibnames=10,
minbibnames=1,
bibstyle=ieee,
]{biblatex}
\AtBeginBibliography{\footnotesize}
\AtEveryBibitem{
  \clearfield{doi}
  \clearfield{isbn}
  \clearfield{issn}
  \clearfield{month}
  \clearfield{publisher}
  \ifentrytype{inproceedings}{
    \clearfield{arxivId}
    \clearfield{archivePrefix}
    \clearfield{eprint}
    \clearfield{url}
  }{}
  \ifentrytype{article}{
    \clearfield{arxivId}
    \clearfield{archivePrefix}
    \clearfield{eprint}
    \clearfield{url}
  }{}
  \ifentrytype{report}{
  }{}
}
\renewbibmacro{in:}{%
  \ifentrytype{inproceedings}{%
    \setunit{}
    \addperiod\addspace In \textit{Proc.\ of the}}%
  {\printtext{\bibstring{in}\intitlepunct}}
}
\DeclareSourcemap{
  \maps{
    \map{ 
      \step[fieldsource=url,
      match=\regexp{\{\\\_\}|\{\_\}|\\\_},
      replace=\regexp{\_}]
    }
    \map{ 
      \step[fieldsource=url,
      match=\regexp{\{\$\\sim\$\}|\{\~\}|\$\\sim\$},
      replace=\regexp{\~}]
    }
    \map{ 
      \step[fieldsource=url,
      match=\regexp{\{\\\x{26}\}},
      replace=\regexp{\x{26}}]
    }
  }
}
\title{\LARGE \bf
D-CLING: Prior-Preserving Depth-Conditioned Fine-Tuning for Navigation Foundation Models
}

\author{Shintaro Nakaoka, 
    Takayuki Kanai, 
    and 
    Kazuhito Tanaka%
\thanks{The authors are with the Frontier Research Center, Toyota Motor Corporation,
Toyota, Aichi, Japan. {\tt \{first\_lastname\}@mail.toyota.co.jp}
}%
}

\begin{document}

\maketitle
\thispagestyle{empty}
\pagestyle{empty}

\begin{abstract}

Navigation Foundation Models (NFMs) trained on large cross-embodied datasets have demonstrated powerful generalizability in various scenarios.
Adopting in-domain fine-tuning for an NFM efficiently calibrates the visuomotor policy, promising further improvement even in a novel scenario.
However, the fine-tuned models still suffer from poor obstacle avoidance or fail to properly reach the provided goals.
Furthermore, model updates using a small subset of data typically erode the pre-trained prior, compromising the pre-training generalization.
Consequently, fine-tuning deteriorates the capability of the model for robust and accurate navigation.
In this work, we present a novel fine-tuning method that leverages large-scale pre-training while efficiently learning in novel setups, such as environments or camera configurations.
In particular, inspired by ControlNet, we fine-tune an NFM by attaching a trainable copy of the pre-trained backbone using zero-initialized residual pathways, thereby learning geometric cues. This design enables the model to efficiently acquire in-domain geometry while preserving pre-trained knowledge across various behaviors.
Despite its simplicity, our comprehensive evaluation of real-world navigation suggests that our proposal effectively enables robust long-horizon navigation with minimal collisions and human intervention. 
Additionally, our offline analysis shows that the proposed method maintains or further improves action prediction capabilities beyond the fine-tuned dataset, providing a key insight into continual learning for general navigation. 
\\The project page:~{\scriptsize \url{https://toyotafrc.github.io/DCLING-Proj/}}

\end{abstract}

\vspace{.5em}
\IEEEoverridecommandlockouts

\section{Introduction}
\label{sec:introduction}

Visual navigation based on a sequence of images has emerged as a fundamental paradigm in mobile robotics research, encompassing diverse tasks~\cite{zhu2017, wijmans2019dd}. 
A key enabler is \emph{imitation learning}, where optimal navigation policies are learned through expert demonstrations in an end-to-end manner~\cite{tai2018socially, ramrakhya2022habitat}.
Navigation Foundation Models (NFMs), such as ViNT~\cite{shah2023vint} and NoMaD~\cite{sridhar2024nomad}, which employ imitation learning based on large-scale demonstrations across various environments and robots, have achieved reliable goal-reaching and obstacle avoidance. 
Importantly, the scale and diversity of the training dataset is critical for acquiring a diverse set of reliable behaviors~\cite{suomela2026data}.
Thus, NFMs typically employ an image-to-action policy learning without additional sensory modalities to enable extensive, low-cost, and standardized pre-training.

\begin{figure}[tb]
	\centering
	\includegraphics[width=\hsize]{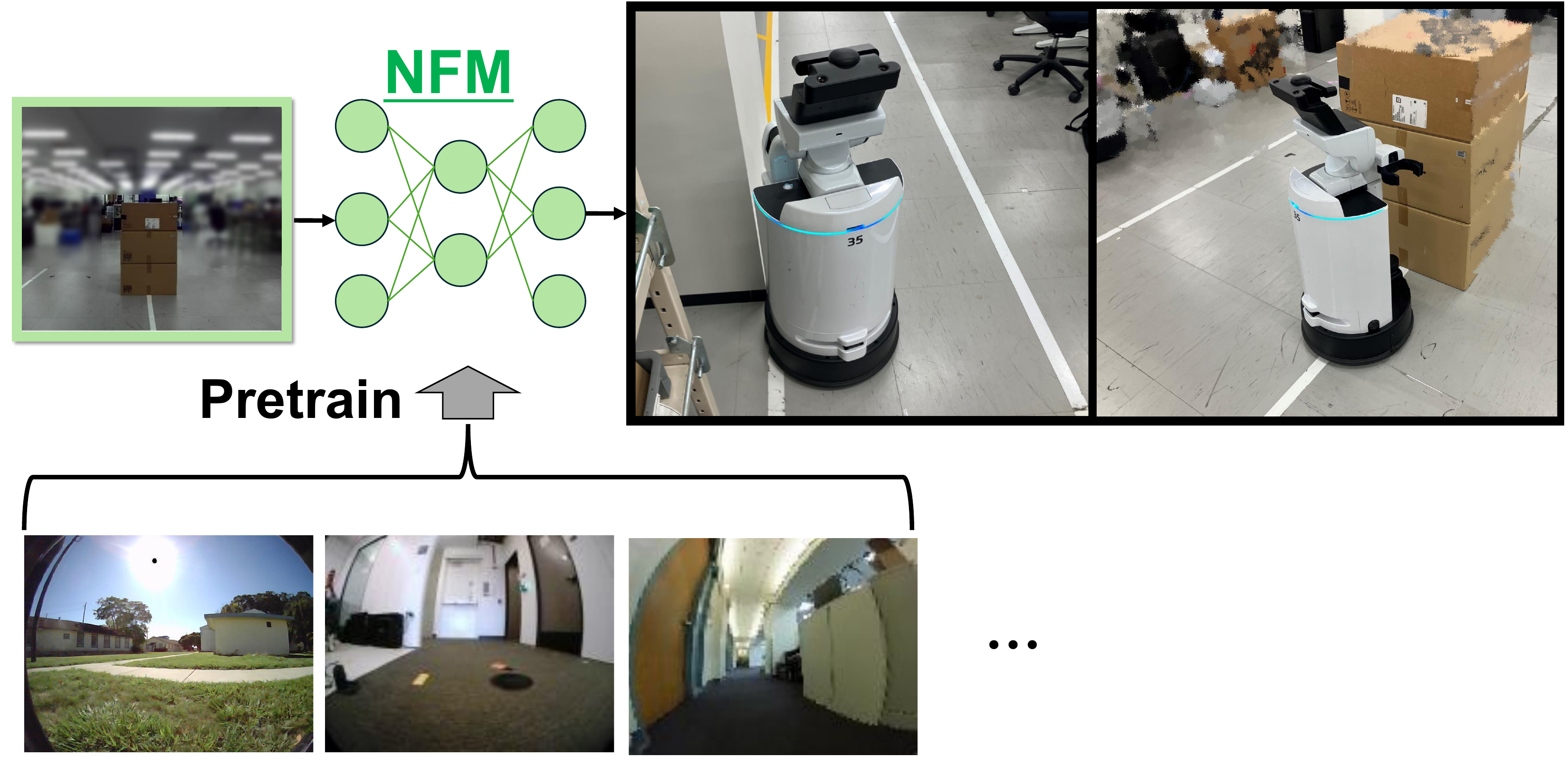}
    \vspace{-1.5em}
	\caption{\textbf{Failure scenarios of the zero-shot NoMaD~\cite{sridhar2024nomad}.} In real-robot navigation, failures including unsafe clearance (left) and distance misestimation (right) were observed.
    In particular, the pre-training images~\cite{hirose2023sacson,shah2021rapid, hirose2019deep, karnan2022socially} exhibited strong distortion relative to the experimental condition, impairing geometric perception.
    }
	\label{fig:failure_case_zero_shot_nomad}
    \vspace{0.5em}
\end{figure}

\begin{figure}[tb]
	\centering
	\includegraphics[width=\hsize]{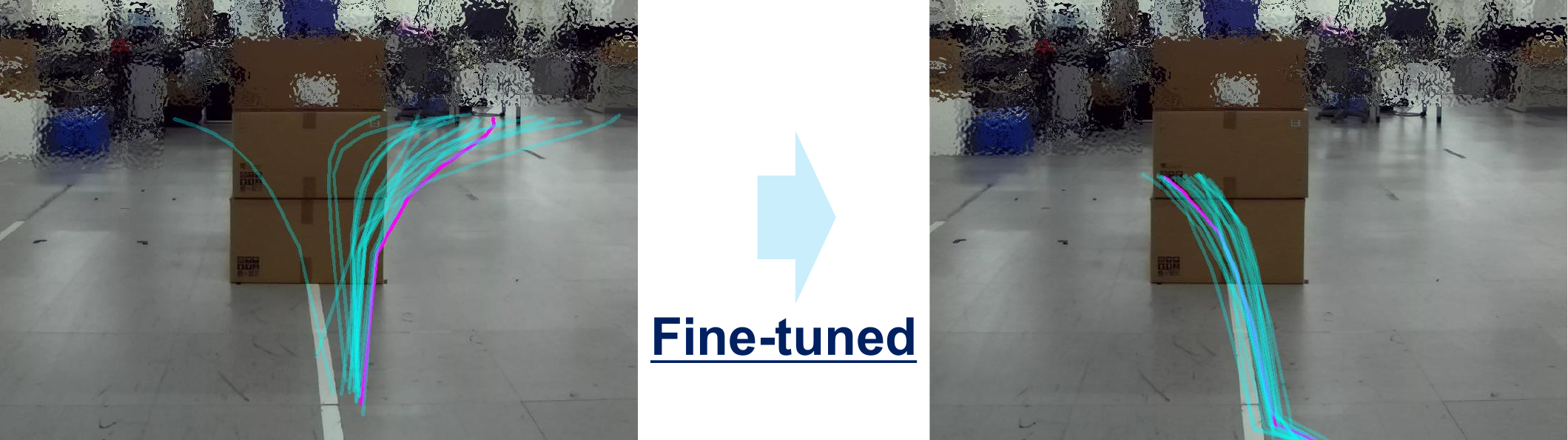}
    \vspace{-1.5em}
	\caption{\textbf{Less divergent trajectory generation after fine-tuning.} Based on a shared start goal and an observation point, \(N=20\) trajectories sampled from each model are shown: \textbf{Left}: the zero-shot \emph{NoMaD} (before fine-tuning); \textbf{Right}: the fine-tuned model. Fine-tuning yielded markedly lower diversity, with a narrower spatial distribution and reduced heading variance, indicating a collapse of the pre-trained priors.}
    \vspace{-1.em}
	\label{fig:waypoint_visualization_before_after_ft}
\end{figure}

However, NFMs struggle to adapt to novel scenes and robotic configurations, even when the scenarios are \textit{seemingly} close to those seen during pre-training.
This is primarily attributed to \textit{domain-shifts} in geometric perception, stemming from differences in camera configurations and/or scene geometry. 
It was observed that providing \textit{more undistorted} images to the NFM than those used for pre-training led to more frequent failures~(Fig.~\ref{fig:failure_case_zero_shot_nomad}).
To address such failures, a common approach is full fine-tuning of the model, that is, calibrating the entire policy backbone to new domains~\cite{shah2023vint,wan2025pig}.
However, it was also observed that the problem persists even after fine-tuning. 
The fine-tuned model is prone to generating less diverse trajectories, suggesting \emph{catastrophic forgetting} of the pre-trained knowledge (Fig.~\ref{fig:waypoint_visualization_before_after_ft}).
In short, techniques for efficiently using large-scale pre-training suffer from two key difficulties: (1) a lack of accurate geometry awareness, given a novel scenario and/or embodiment; and (2) insufficient behavioral diversity, which is crucial for dealing with the various navigational scenarios. 

\begin{figure}[t]
	\centering
	\includegraphics[width=\hsize]{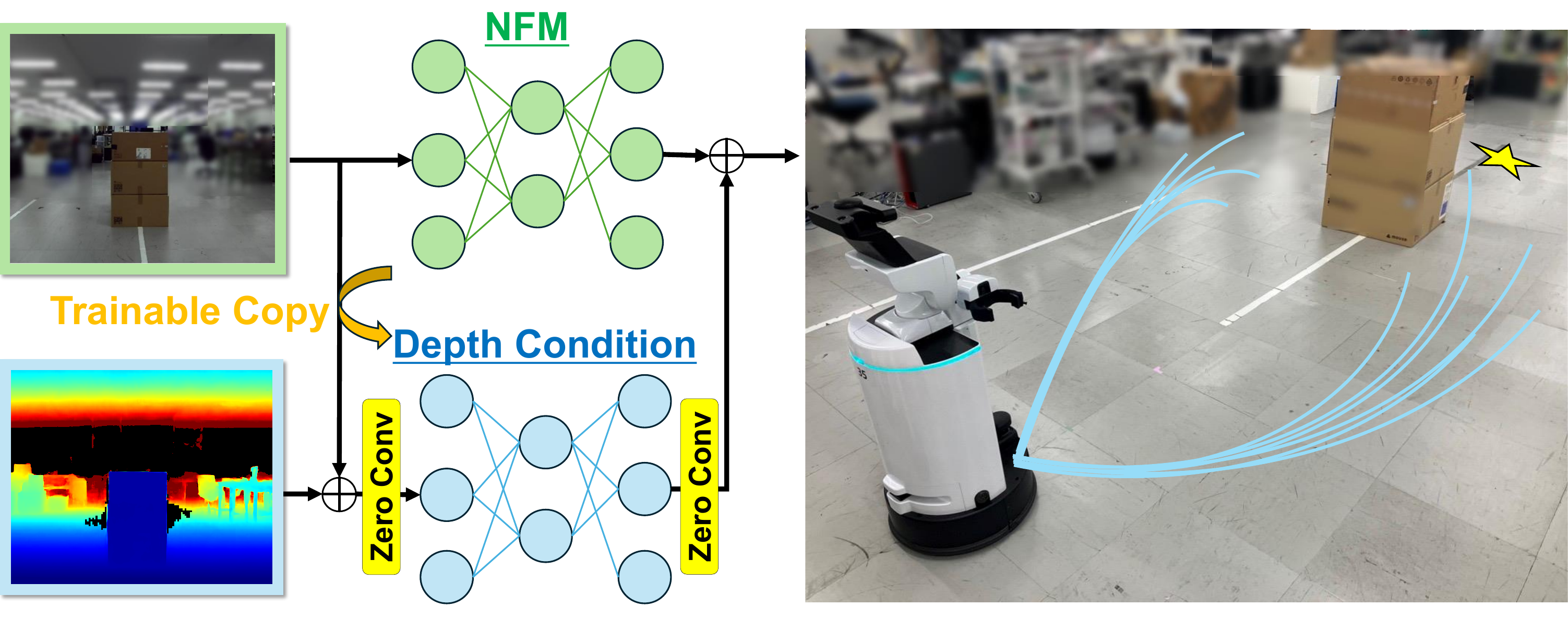}
	\vspace{-5pt}
	\caption{\textbf{Overview of the proposed method}: prior-preserving dense-depth conditioning, inspired by ControlNet~\cite{zhang2023adding}. Zero-initialized residual pathways inject per-pixel depth into intermediate features, preserving the pre-trained prior and improving geometry-consistent obstacle avoidance across cameras.}
    \vspace{-1.0em}
	\label{fig:overview_of_our_approach}
\end{figure}

\newcommand{\dclingOrig}{
    We named D-CLING, hoping the model \emph{\textbf{cling}} the pre-trained knowledge of NFMs without \emph{catastrophic forgetting} it in the \emph{\textbf{d}epth}-guided tuning. 
}

This study presents a novel fine-tuning method for NFMs, \textbf{D\mbox{-}CLING}\footnote{\dclingOrig} (\textit{\textbf{D}epth\mbox{-}conditioned, \textbf{C}ontrolNet\mbox{-}driven \textbf{L}earn\textbf{I}ng for General \textbf{N}avi\textbf{G}ation Models}), 
that leverages large\mbox{-}scale pre-training while efficiently adapting to novel environments and geometric perception (Fig.~\ref{fig:overview_of_our_approach}). 
The main concept is the fine-tuning a policy backbone with \textit{dense-depth} conditioning to capture accurate scenario geometry, using ControlNet\mbox{-}style residual pathway learning~\cite{zhang2023adding}.
The pathway injects dense depth signals into the intermediate layers of the model, progressively updating the parameters with geometry-awareness.
Hence, the fine-tuned model is expected to preserve the original navigation capability while smoothly increasing its in-domain geometry-awareness, resulting in robust and accurate navigation after the fine-tuning.

This study evaluates D-CLING built upon NoMaD~\cite{sridhar2024nomad}, a standard NFM pre-trained on diverse domains.
Real-world evaluation demonstrates that D-CLING offers a substantial improvement in goal-reachability and collision avoidance skills relative to the baselines: the zero-shot NoMaD, RGB fine-tuned models, and RGB-D fine-tuned models in a common (\emph{early-fusion}) strategy~\cite{gode2024flownav}. 
Additionally, offline analysis of both the fine-tuned model and its pre-trained domain shows that D-CLING not only adapts to the former \emph{but remains} powerful in the latter, indicating preservation and extension of the pre-trained prior.

In summary, the main contributions of this study are as follows:
\begin{itemize} 
	\item \textbf{Prior-preserving framework for fine-tuning of the navigation model:} A depth-conditioned adaptation is introduced that retains pre-trained policy priors while explicitly injecting geometry-awareness.

     \item \textbf{Comprehensive validation:} Comprehensive experiments are presented, showing that D-CLING achieves superior goal reachability and obstacle avoidance in both real-robot deployments and offline evaluations compared with typical baselines.

    \item \textbf{Experimental evidence for future extension:} 
    The proposed fine-tuning method further extends navigation performance beyond the fine-tuned domain, suggesting the potential for future extensions of NFMs.

\end{itemize}

\section{Related Work}
\subsection{Visual Navigation}
Traditional visual navigation pipelines are often modular~\cite{thoma2019mapping}, consisting of (1) mapping and localization~\cite{mur2015orb};
(2) visual perception, such as terrain traversability or object-category estimation~\cite{Gasparino2022};
and (3) path planning and low-level control.
While modular strategies perform well in relatively simple environments~\cite{chaplot2020learning}, the hand-crafted interfaces between modules may discard critical cues, thereby reducing the robustness of the overall system.

In contrast, recently developed end-to-end strategies, such as reinforcement learning (RL)~\cite{stachowicz2024lifelong} and imitation learning~\cite{sridhar2024nomad}, overcome difficulties by acquiring critical queues from data with minimal assumptions. 
In particular, imitation learning has gained significant traction because it avoids time-consuming reward engineering~\cite{qiu2020learning} and offers a higher sample efficiency~\cite{zhu2017}. 

In addition, it can train the navigation policy directly using real-world teleoperation logs,
without relying on long-horizon exploration. 
Accordingly, imitation-learning–based navigation has been widely applied across diverse navigation tasks, including PointGoal~\cite{zhang2025creste, liu2025citywalker}, ImageGoal~\cite{wan2025pig, feng2025image}, Object Navigation, Vision-and-Language Navigation~\cite{anderson2018vision, kamath2023new, hirose2024lelan}, and Social Navigation~\cite{hirose2023sacson, song2024vlm}.

Despite its practical advantages, imitation-learning-based navigation remains vulnerable to domain shift in novel domains, such as changes in layout/illumination~\cite{ijcai2020p124} or camera extrinsics/intrinsics~\cite{tang2022mono}, which can substantially reduce the task success.
To overcome these limitations, recent work has introduced NFMs~\cite{sridhar2024nomad,liu2025citywalker,cai2025navdp,wang2025x}, which are end-to-end-trained models on large heterogeneous datasets spanning multiple robots and environments. 
Relying only on RGB lowers hardware and data-collection costs, avoids multi-sensor calibration and synchronization, and maintains the comparability of datasets across platforms. 
This study is based on NFMs, which have been further extended by using small-scale training whilst avoiding the need for additional large-scale training.

\subsection{Fine-tuning for Navigation Foundation Model}

\begin{figure*}[!t]
	\centering
	\includegraphics[width=0.88\hsize]{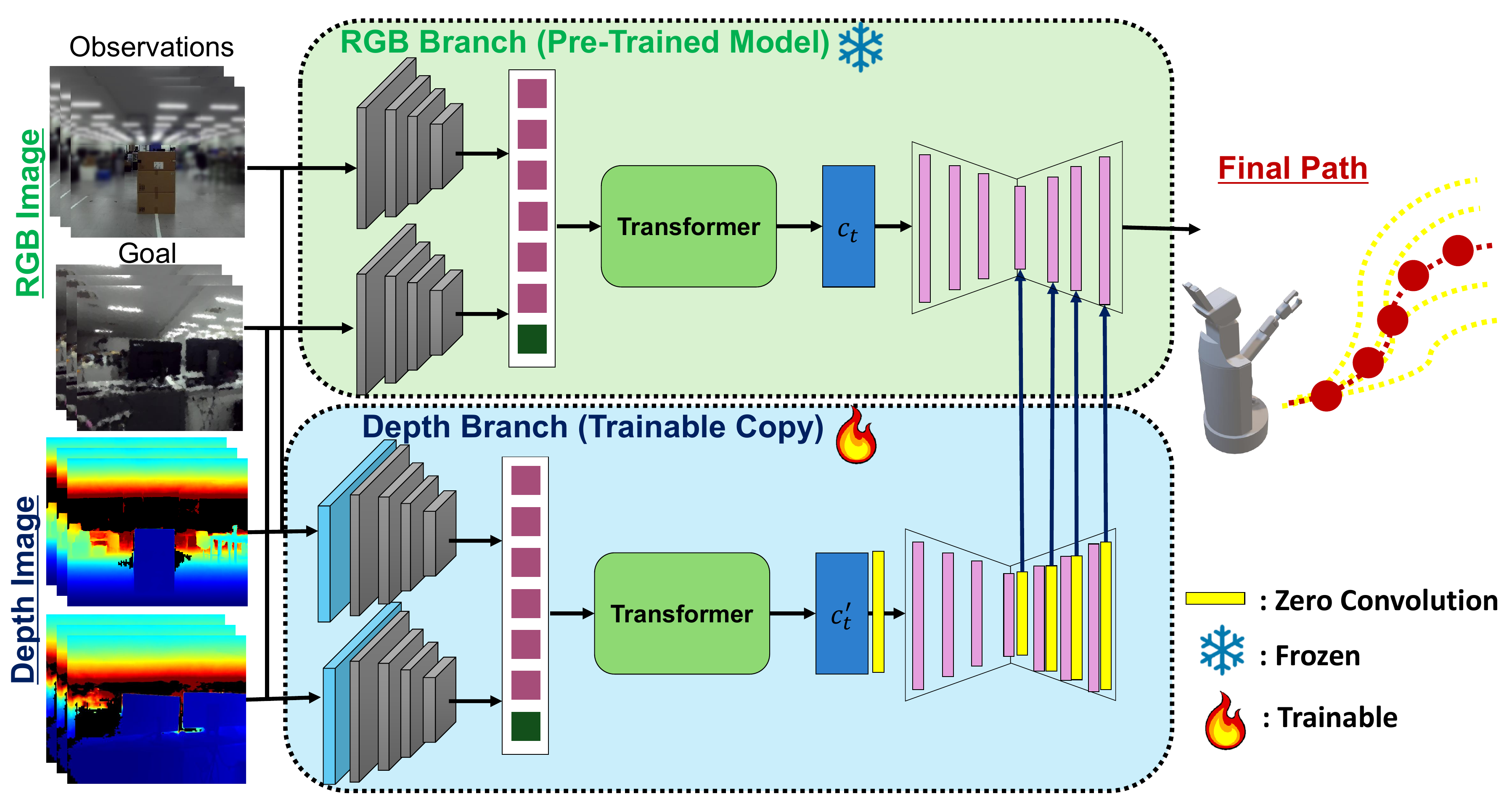}
	\vspace{-8pt}
    	\caption{\textbf{D-CLING architectural design, based on NoMaD~\cite{sridhar2024nomad}.} A frozen pre-trained \emph{RGB Branch} (identical to the NoMaD architecture) maps an RGB history and goal image to intermediate features. In parallel, a trainable \emph{Depth Branch} ingests RGB–D (with a $4\!\to\!3$ embedding) and injects zero-initialized residual features into the U-Net-based diffusion model; the two streams are fused by element-wise addition at each U-Net stage during training and inference. The diffusion head outputs a short-horizon action distribution, enabling prior-preserving depth conditioning.
	}
	\label{fig:model_architecture}
\end{figure*}

Fine-tuning is a common practice for further extending the capabilities of zero-shot models, as domain-specific knowledge is a key driver of task performance~\cite{shah2023vint}.
This strategy offers a strong sample efficiency, enabling rapid adaptation from few demonstrations~\cite{barreiros2025careful}.
To fine-tune the visuomotor policy, a standard strategy prioritizes downsizing and maintaining the \emph{same I/O}. As the model typically (1) possesses large-scale parameters to handle a large amount of data and (2) can suffer from catastrophic forgetting of the pre-trained knowledge, such strategies are promising for feasible yet efficient real-robot adaptation.
Based on the powerful zero-shot models~\cite{o2024open, intelligence2025pi_, bjorck2025gr00t}, parameter-efficient fine-tuning (PEFT), such as LoRA and adding adapter layers, has been shown to preserve pre-trained priors while enabling lightweight adaptation in manipulation settings, mitigating the forgetting and overfitting of priors~\cite{  Liu2024,lu2024learning,sharmalossless,bousmalis2023robocat}.

In contrast, fine-tuning of NFMs need not always prioritize downsizing, as their parameter scale is typically negligible.
This is because the navigation task must be performed in \emph{real-time}, and thus, the models are designed in a lightweight manner even for NFMs~\cite{shah2023vint,sridhar2024nomad}.
This property of NFMs addresses the ‘off-the-shelf’ use of the large-scale backbone, and/or running in parallel for fine-tuning and inference.
Therefore, this study employs ControlNet-like fine-tuning as a key strategy~\cite{zhang2023adding}, which prioritizes \emph{preserving} the diversity of data generation patterns over the memory footprint and inference speed of the model.
Moreover, the ControlNet-based update of the backbone is localized rather than updating the entire backbone, which is less prone to overfitting, even with a small-scale dataset~\cite{Liu2024,lu2024learning,sharmalossless}.  
Therefore, policy tuning can be achieved without a costly dataset while maintaining the pre-trained knowledge.
A closely related study explores lifelong autonomous improvement of NFMs in the wild by combining offline RL pre-training with autonomous online RL fine-tuning~\cite{stachowicz2024lifelong}.
Without relying on real-world exploration or an RGB-D pre-trained NFM backbone, this study shows that the ControlNet-inspired strategy, D-CLING, efficiently improves the robustness of the NFM policy under human-controlled demonstrations.
Moreover, this study experimentally quantifies that the proposed method is more suitable for preserving the pre-trained knowledge than the naive RGB-D fusion strategy, that is, the \emph{early-fusion}~\cite{gode2024flownav}, 
indicating the insight for the future fine-tuning design beyond this work.
\section{Problem Formulation}
\label{sec:proposed_method}
The objective of this study is to learn a policy $\pi_{\theta}$ that:
\begin{equation}
\pi_{\theta}:
\bigl(\tilde{\mathbf{o}}_{t-T:t},\,\mathbf{g}\bigr)
\;\longmapsto\;
p_{\theta}\!\left(\mathbf{a}_{t:t+H}\,\middle|\,\tilde{\mathbf{o}}_{t-T:t},\,\mathbf{g}\right),
\end{equation}

\noindent
where a sequence of observations $\tilde{\mathbf{o}}_{t-T:t}=(\tilde{\mathbf{o}}_{t-T},\dots,\tilde{\mathbf{o}}_{t})$ and a goal image $\mathbf{g}$ provide a probability distribution $p_{\theta}(\cdot)$ over future action sequences $\mathbf{a}_{t:t+H}=(\mathbf{a}_{t},\dots,\mathbf{a}_{t+H})$, thereby enabling a mobile robot to reach its goal using only visual observations.

Here, $\tilde{\mathbf{o}}_t \in \mathbb{R}^{h \times w \times 4}$ denotes a single RGB-D observation obtained by concatenation of the raw RGB image with the depth map, assuming that this is provided by a learning-based depth-estimation network. The goal information is provided as a reference RGB image $\mathbf{g} \in \mathbb{R}^{h \times w \times 3}$ depicting the target viewpoint.
$h$ and $w$ denote the image height and width, respectively, and $\mathbf{a}_t \in \mathbb{R}^{2}$ is a 2-D action vector that represents a waypoint in the current robot frame at time $t$. The hyper-parameter $H$ denotes the prediction horizon, while $T$ is the number of past frames used as input.

\section{Methodology}

\subsection{Preliminaries: NoMaD~\cite{sridhar2024nomad}}
\label{sec:nomad}

NoMaD~\cite{sridhar2024nomad} is illustrated in the upper part of Fig.~\ref{fig:model_architecture}.
The model receives the current RGB frame together with the $T$-frame RGB history
$\mathbf{o}_{t}=o_{t-T:t},$
and predicts a distribution over the action sequence from the current step through $H$ future steps $\mathbf{a}_{t}=a_{t:t+H}$.
Visual features are extracted with the Visual Navigation Transformer (ViNT)~\cite{shah2023vint}.
The current frame $o_{t}$ and the goal image $o_{g}$ are embedded by an encoder $\phi(\cdot)$,
whereas the history window $o_{t-T:t}$ is embedded by a separate encoder $\psi(\cdot)$.
The resulting vectors are fed into a Transformer block $f_{\mathrm{tr}}(\cdot)$,
yielding the context vector $\mathbf{c}_t$ as follows:

\begin{equation} \mathbf{c}_t=f_{\mathrm{tr}}\!\bigl(\psi(o_{t-T:t}),\,\phi(o_{t},o_{g})\bigr). \end{equation}

Conditioned on $\mathbf{c}_t$, a diffusion model with parameters $\theta$
defines the conditional distribution
$p_{\theta}\!\left(\mathbf{a}_{t:t+H}\mid \mathbf{c}_t\right)$,
from which a sample is drawn:
\begin{equation}
\hat{\mathbf{a}}_{t:t+H} \sim p_{\theta}\!\left(\cdot \mid \mathbf{c}_t\right).
\end{equation}

\subsection{Overview of the Proposed Method}
\label{sec:Overview_proposed_method}

Figure~\ref{fig:model_architecture} presents the proposed framework adopted to a representative NFM, NoMaD~\cite{sridhar2024nomad}.
As in the original NoMaD, the pre-trained weight-frozen NoMaD backbone (\textit{RGB Branch}) maps a short RGB history and a goal image to actions. In parallel, a depth-conditioned branch (\textit{Depth Branch}) encodes the same RGB inputs with a depth map to produce conditioning features. The model then outputs short-horizon waypoints.

\subsection{Model Architecture}
\label{sec:depth_conditioned_branch}
All layers of the pre-trained NoMaD (\textit{RGB Branch}) are frozen and a copy is created to form the \textit{Depth Branch}.
The \textit{Depth Branch} receives an RGB-D frame
 $\tilde{\mathbf{o}}_{t}\in\mathbb{R}^{h\times w\times 4}$ and begins with a $4\!\rightarrow\!3$ embedding layer that projects the four-channel input to three channels.
All subsequent modules follow NoMaD.
Conditioned on the context vector $c_t'$, a U-Net-based diffusion model of the \textit{Depth Branch} produces intermediate features.
At every corresponding U-Net layer, depth intermediate features are added to the \textit{RGB Branch}. 

Following ControlNet, zero-initialized $1{\times}1$ convolutions are inserted immediately before the U-Net and immediately after each U-Net layer on the \textit{Depth Branch}.
Let \(F_{\ell}(\cdot;\Theta_{\ell})\) denote the intermediate block at stage \(\ell \in \{1,\dots,L\}\) of the U-Net-based diffusion model of the \textit{RGB Branch}, with an input feature \(h_{\ell}\) and output \(y_{\ell} = F_{\ell}(h_{\ell};\Theta_{\ell})\).
Here, \(\Theta_{\ell}\) denotes the model parameters of \(F_{\ell}\). In the \textit{RGB Branch}, the parameters $\Theta_{\ell}$ of $F_{\ell}$ are frozen.

For the \textit{Depth Branch}, a counterpart
\(F_{\ell}^{d}(\cdot;\Theta_{\ell}^{d})\) and a single zero-initialized \(1{\times}1\)
convolution \(Z_{\ell}(\cdot;\Theta_{\ell}^{z})\) are introduced.
With the depth-derived feature \(h_{\ell}^{d}\), let
\(u_{\ell}^{d} = F_{\ell}^{d}(h_{\ell}^{d};\Theta_{\ell}^{d})\) be the
intermediate feature of the U-Net based diffusion model at stage \(\ell\).
The block output is formed as the element-wise sum of \(y_{\ell}\) and the
zero-initialized \(1{\times}1\) convolution applied to \(u_{\ell}^{d}\):
\begin{equation}
\label{eq:controlnet-inject}
y'_{\ell}
= y_{\ell} + Z_{\ell}~\!\bigl(
     u_{\ell}^{d}
     \,;\,\Theta_{\ell}^{z}
\bigr).
\end{equation}

\noindent
Importantly, the \(1{\times}1\) fusion gate is zero-initialized as follows:
\begin{equation}
\Theta_{\ell}^{z}=\mathbf{0}
\ \Longrightarrow\
\forall x:\ Z_{\ell}(x;\Theta_{\ell}^{z})=\mathbf{0}
\end{equation}
Hence, at the initialization phase, all the layer-wise outputs $y_{\ell}$ behave as their original form, that is, \(y'_{\ell}=y_{\ell}=F_{\ell}(h_{\ell};\Theta_{\ell})\).
Thus, the gradients update the \textit{Depth Branch} parameters gradually via the zero-initialized fusion, while the RGB trunk remains frozen.

This approach is a relatively simple adoption of the ControlNet philosophy to validate the impact of the proposed method. 
Although further extensions, such as a repulsive safety head from monocular depth~\cite{kim2025enhancing} and externally providing a 3-D map~\cite{honda2025gsplatvnm}, can be integrated for future extensions, they are intentionally excluded in this study.

\section{Experiments}
\label{sec:experiments}
\begin{table*}[t]
  \centering
  \small
  \setlength{\tabcolsep}{10pt}  
  \renewcommand{\arraystretch}{1.2}
  \caption{\textbf{Real-world navigation performance} across three scenarios. 
  The average success rate (SR) in $10$ trials each for scenarios (i) and (ii), and the average human interventions (interventions) of $5$ trials for scenario (iii) are listed.
  }
  \label{tab:real_world_results}
  \begin{tabular}{lccccc}
    \toprule
    \multirow{2}{*}{Method} & \multirow{2}{*}{Training} & \multirow{2}{*}{Modality}
      & (i) \textit{Basic Obstacle} & (ii) \textit{Dynamic Corridor} & (iii) \textit{Long-range} \\
    \cmidrule(lr){4-4}\cmidrule(lr){5-5}\cmidrule(lr){6-6}
    & & & SR (\%) $\uparrow$ & SR (\%) $\uparrow$ & Interventions $\downarrow$ \\
    \midrule
    NoMaD~\cite{sridhar2024nomad} & Frozen         & RGB    & \runup{50} & 0 & \runup{2.6} \\
    NoMaD-FT                      & Full fine-tune & RGB    & 30 & \runup{10} & 3.2 \\
    NoMaD-EF                      & Early fusion   & RGB-D  & 40 & 0 & 4.4 \\
    \midrule
    \textbf{D-CLING (Ours)}      & \textbf{Zero-init} & \textbf{RGB-D} & \textbf{70} & \textbf{60} & \textbf{1.2} \\
    \bottomrule
  \end{tabular}
\end{table*}

\begin{figure*}[tb]
  \centering

  \begin{minipage}[t]{0.32\linewidth}
    \centering
    \includegraphics[width=\linewidth,keepaspectratio]{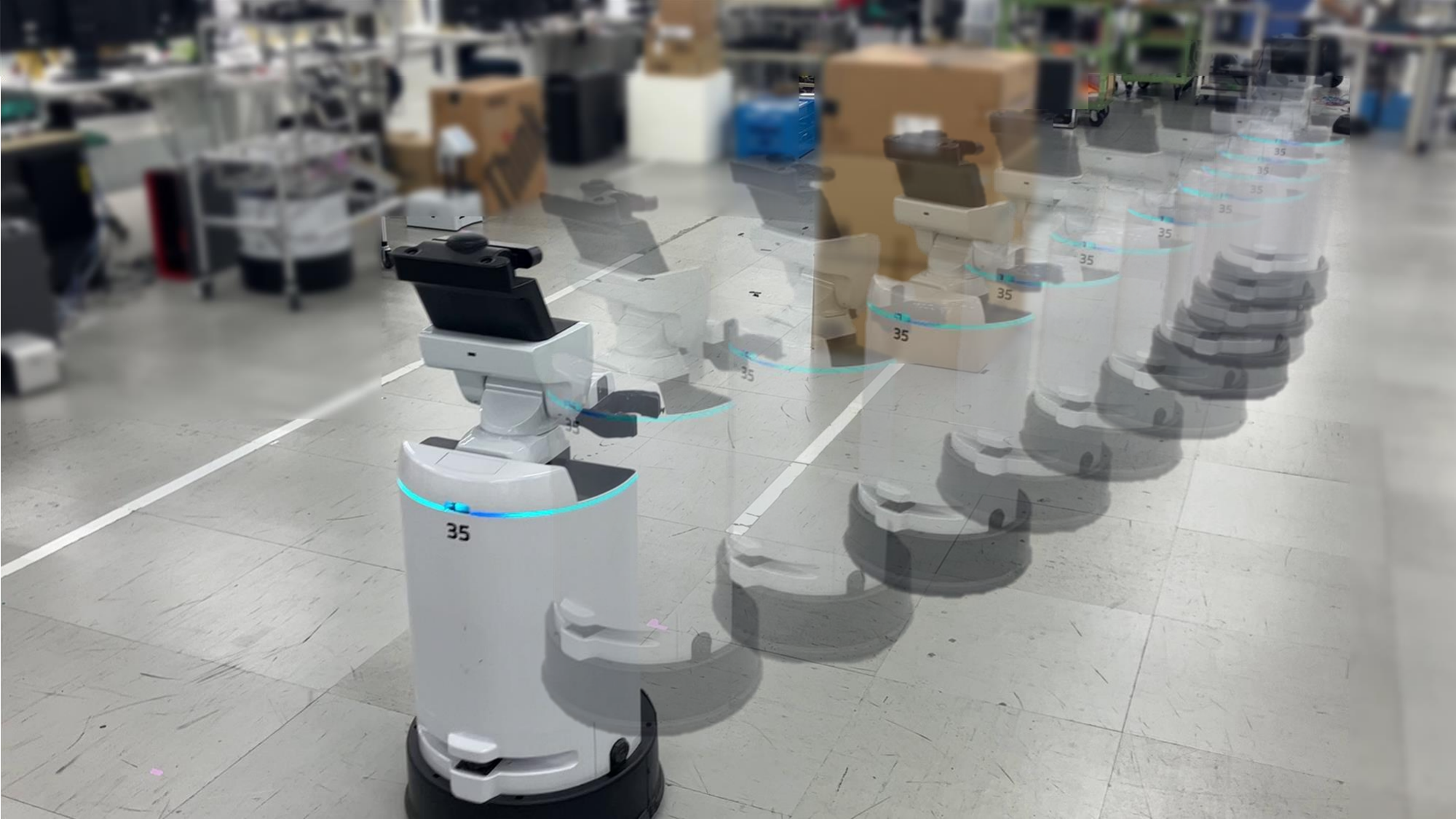}
  \end{minipage}\hfill
  \begin{minipage}[t]{0.32\linewidth}
    \centering
    \includegraphics[width=\linewidth,keepaspectratio]{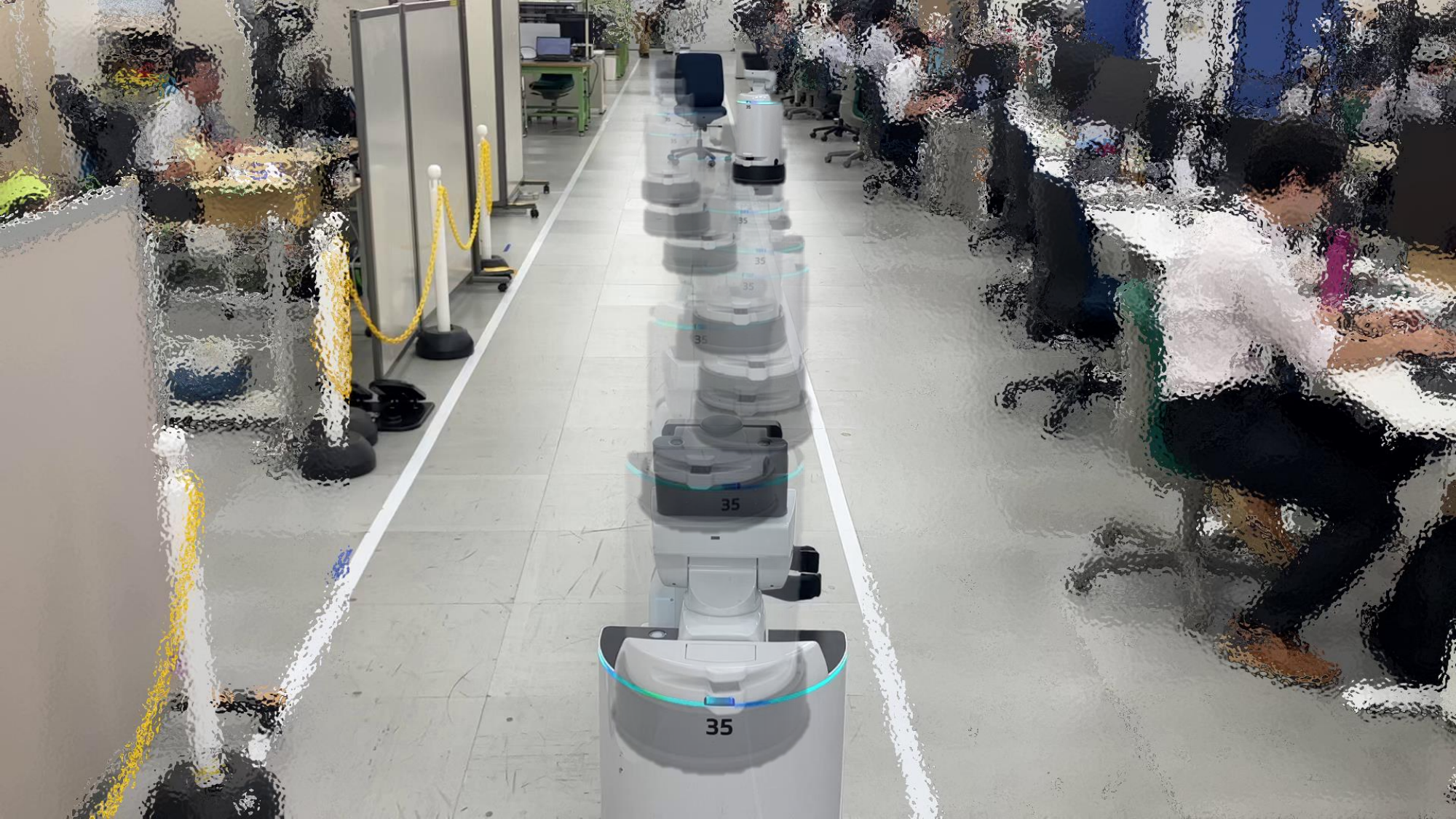}
  \end{minipage}\hfill
  \begin{minipage}[t]{0.32\linewidth}
    \centering
    \includegraphics[width=\linewidth,keepaspectratio]{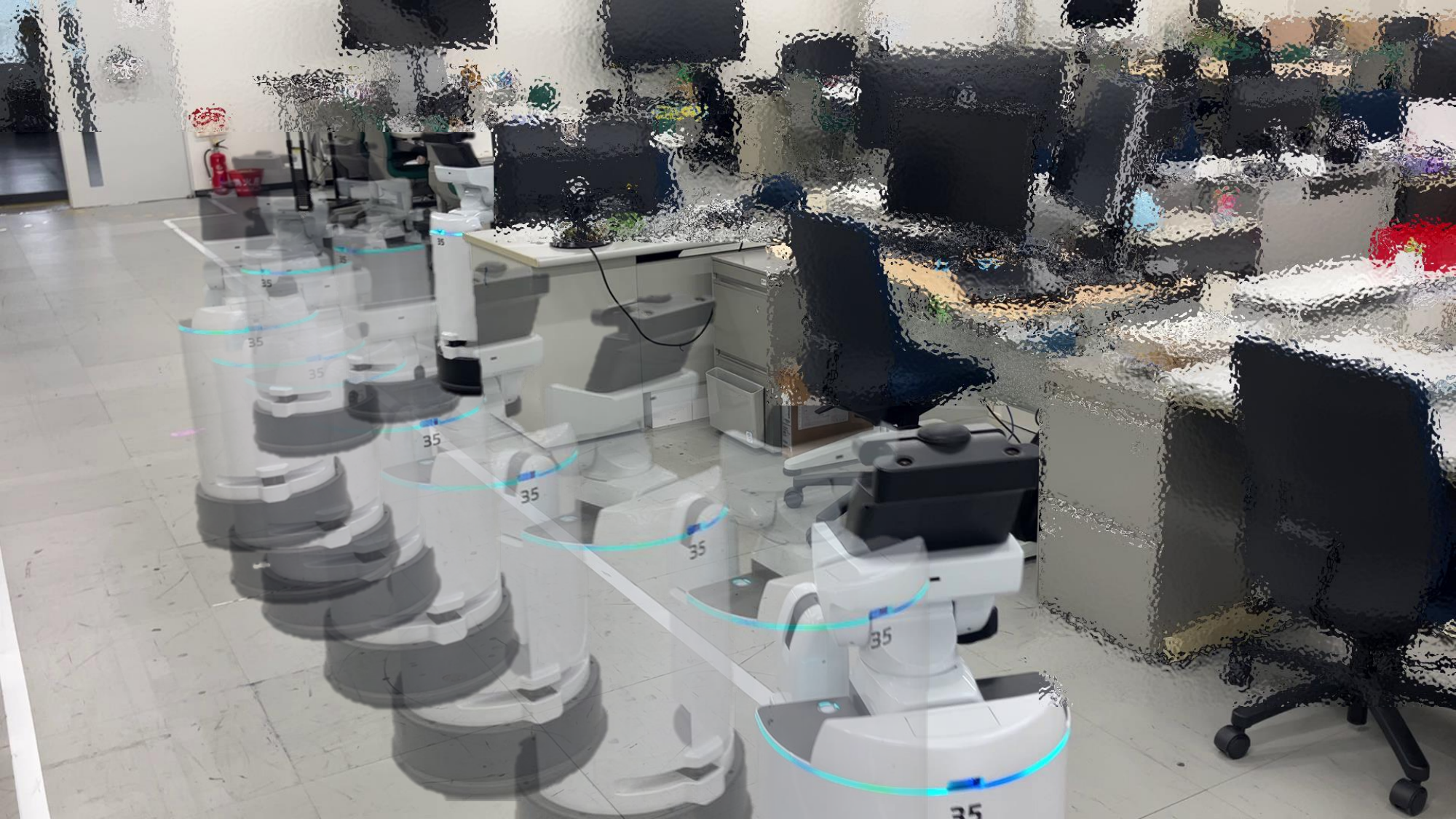}
  \end{minipage}

  \begin{minipage}[t]{0.32\linewidth}
    \centering
    \includegraphics[width=\linewidth,keepaspectratio]{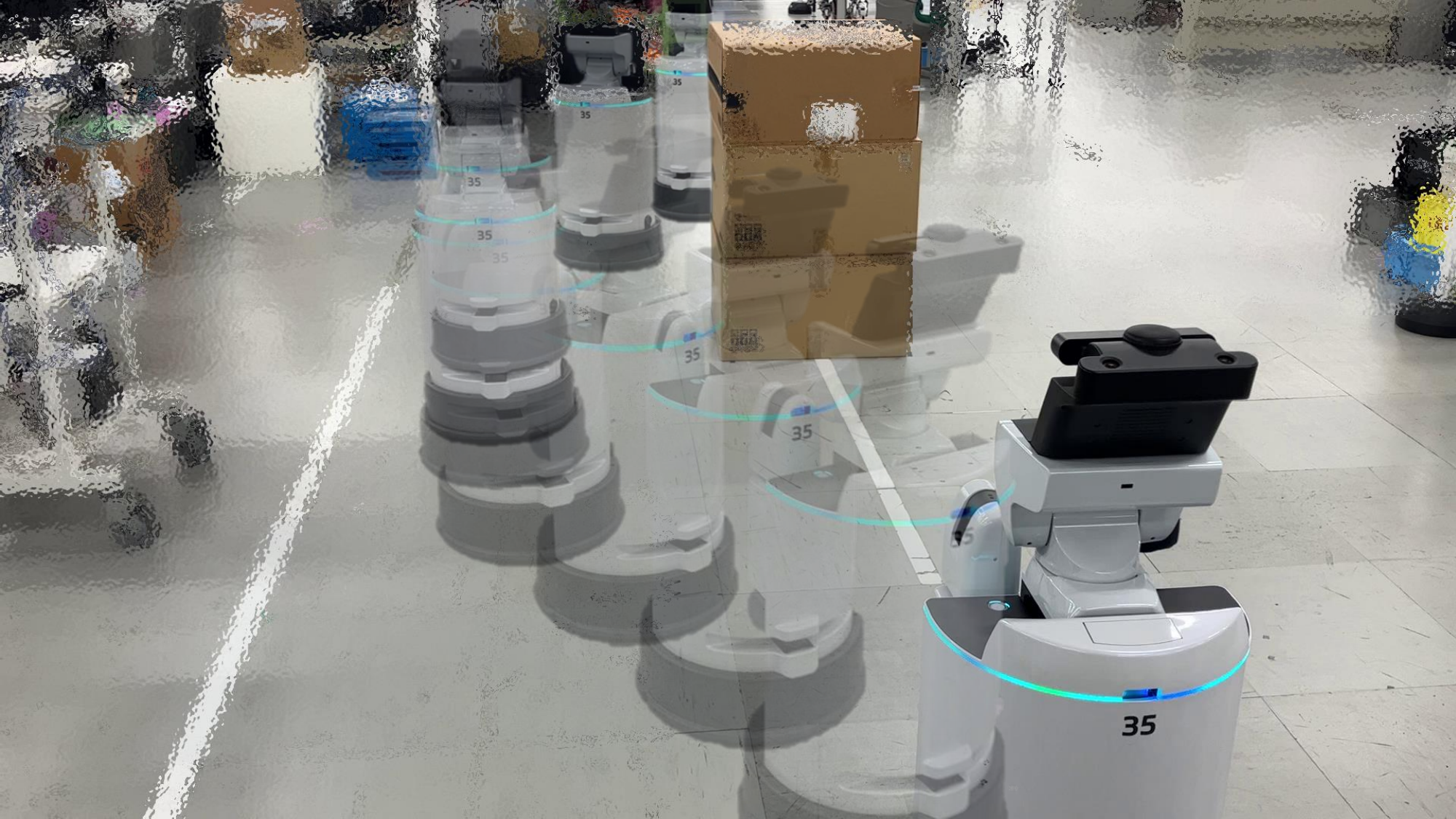}
    \vspace{0.3em}
    {\footnotesize (i) \textit{Basic Obstacle}}
  \end{minipage}\hfill
  \begin{minipage}[t]{0.32\linewidth}
    \centering
    \includegraphics[width=\linewidth,keepaspectratio]{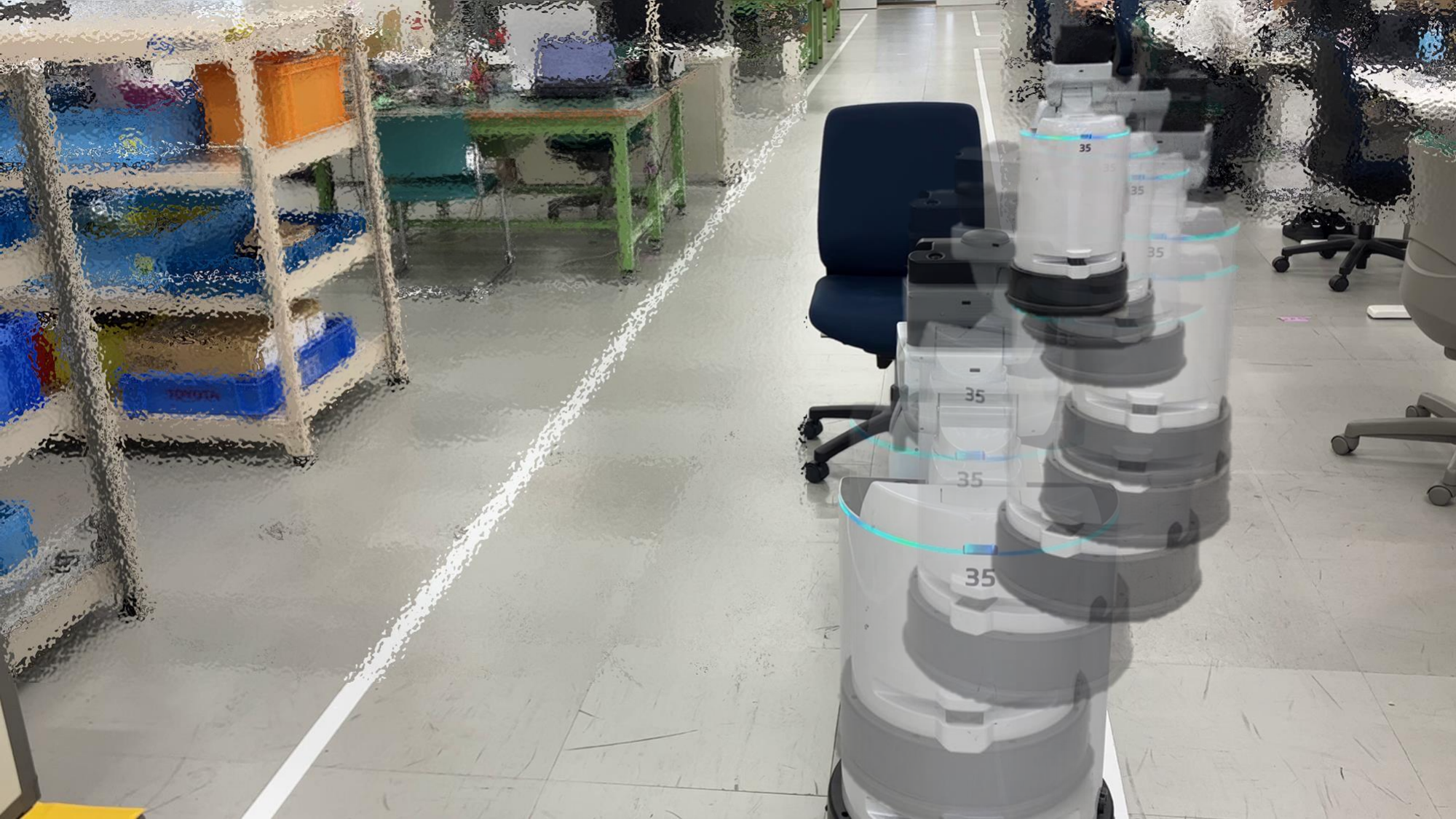}
    \vspace{0.3em}
    {\footnotesize (ii) \textit{Dynamic Corridor}}
  \end{minipage}\hfill
  \begin{minipage}[t]{0.32\linewidth}
    \centering
    \includegraphics[width=\linewidth,keepaspectratio]{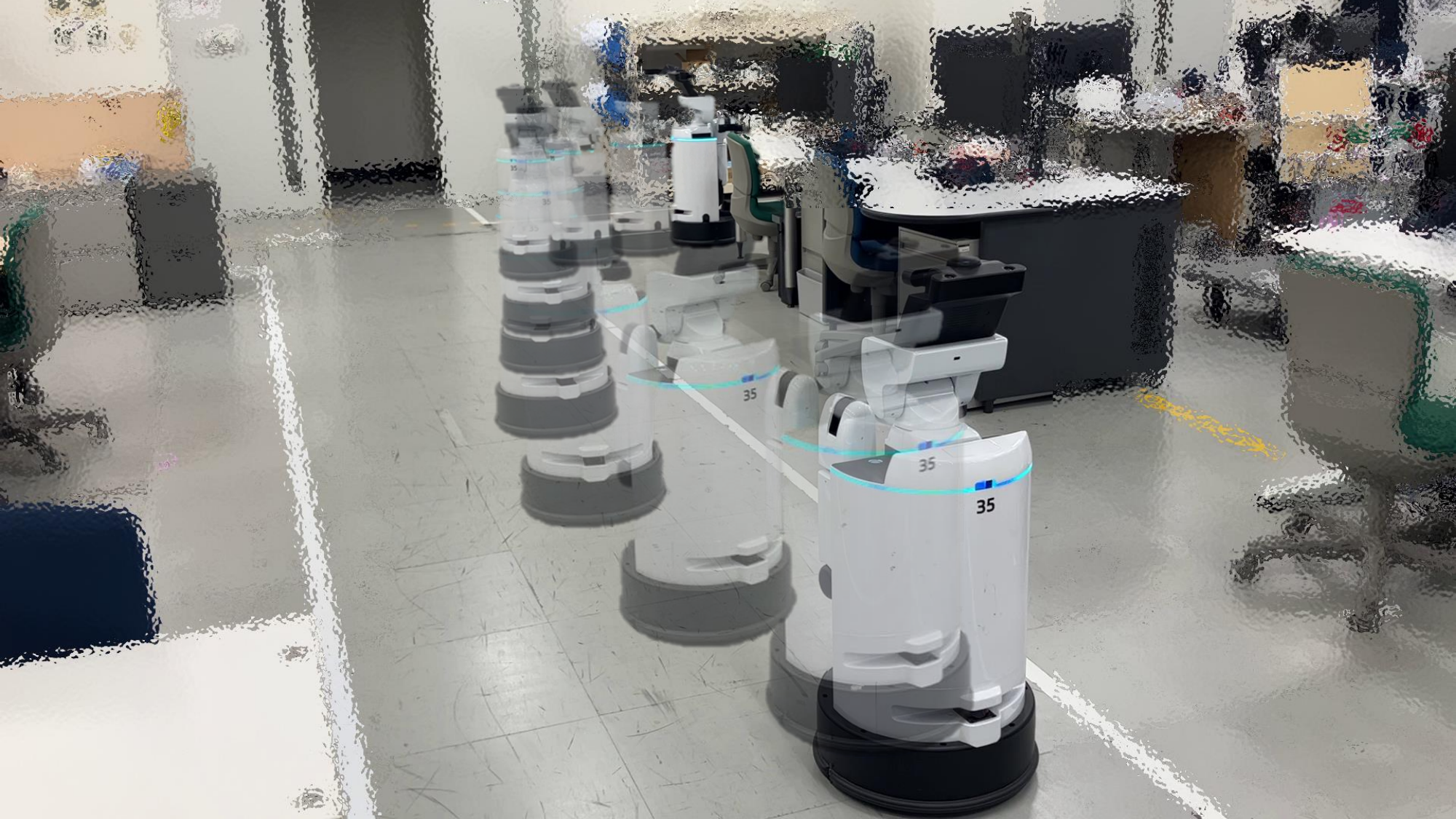}
    \vspace{0.3em}
    {\footnotesize (iii) \textit{Long-range}}
  \end{minipage}

  \caption{\textbf{Representative frames of the proposed method from real-world experiments with a physical overlaid robot in an office environment:} (i) \emph{Basic Obstacle}—corridor traversal with visual avoidance of a single stationary box; (ii) \emph{Dynamic Corridor}—after $10\,\mathrm{m}$ the robot must avoid an unmapped chair; and (iii) \emph{Long-range}—a $50\,\mathrm{m}$ semicircular route across two junctions.}
  \label{fig:real_world_experiment}
\end{figure*} 

\subsection{Fine-tuning Setups}
\label{sec:experiments_setup}

\noindent \textbf{Dataset construction.}
We collected synchronized RGB-odometry sequences using a Toyota Human Support Robot (HSR)~\cite{Yamamoto2019} equipped with a ZED~2. 
A learning-based stereo-to-depth estimator pre-trained on in-house datasets~\cite{shankar2022learned} was used for dense depth estimation.
The sequences are collected in a large-scale office room. The space combines standard office furniture with specialized robotics equipment and experimental setups, resulting in a heterogeneous environment that challenges navigation with both typical and atypical obstacles.
The dataset also contains scenarios requiring avoidance of dynamic obstacles, in addition to static obstacles.
The demonstration data for the dedicated fine-tuning dataset, RealHSRNav, was collected in approximately 3 h. 
RealHSRNav was used as the \emph{sole} dataset for all fine-tuning experiments in this study.

\noindent \textbf{Implementation details.}
To implement D-CLING, we fine-tuned an off-the-shelf checkpoint of the NoMaD\footnote{\label{fn:nomad}\url{https://github.com/robodhruv/visualnav-transformer} (retrieved 10 July 2025)} for $30$~epochs on a single NVIDIA RTX\,4090\,GPU with a batch size of~256 and a learning rate of $2.5\times10^{-5}$.
Following the approach of ~\cite{sridhar2024nomad},
we train with AdamW~\cite{loshchilov2017decoupled} using a cosine learning-rate schedule with warm-up, optimizing the parameters with respect to the unmodified NoMaD loss function.

Importantly, the selection of the checkpoint and RealHSRNav dataset poses a domain-shift, particularly due to differences in the camera field of view. Although the model was originally pre-trained on \emph{fisheye}-like images, mostly, the equipped camera provides a \emph{pinhole}-like projection (approximately $110^\circ$ horizontal).
Thus, adequately \emph{calibrating} scene perception is needed to leverage the pre-trained knowledge.

\subsection{Baselines}
\label{subsec:baselines}

We used the following ablative models to compare with the proposed D-CLING:

\noindent
\textbf{NoMaD (Zero-Shot).} We evaluate the same NoMaD checkpoint for the fine-tuning variants, including our proposal.
We show that this zero-shot adoption frequently achieves a \textit{runner-up} position in various tasks, supported by the knowledge obtained in large-scale pre-training.

\noindent
\textbf{NoMaD-FT (Full Fine-Tuning).} All NoMaD parameters are fine-tuned on our dataset under identical conditions to D-CLING.
This baseline shows the difficulty of in-domain learning: naive fine-tuning of the zero-shot model is insufficient, and, in fact, degrades its original ability.

\noindent
\textbf{NoMaD-EF (Early Fusion).} Based on the NoMaD checkpoint, we added a depth encoder with the same architecture as the RGB encoder, which is trainable and randomly initialized. 
This is a common practice to adopt multi-modal model input~\cite{gode2024flownav}: a trainable depth-only backbone is added parallel to the RGB backbone and tokens are fused via channel-wise concatenation followed by a \(1{\times}1\) projection without additional residual paths.

\subsection{Real-world Experiments}
\label{sec:real_world_experiments}

\noindent\textbf{Scenarios.}
We evaluate all the methods in the following three scenarios, which align with real-world scenarios of navigation tasks (Fig.~\ref{fig:real_world_experiment}):

\begin{enumerate}[label=(\roman*)]
    \item \textbf{Basic obstacle avoidance (\emph{Basic Obstacle})}. The robot traverses a corridor while avoiding a stationary box; no additional obstacles are introduced during the evaluation.
    \item \textbf{Dynamic corridor (\emph{Dynamic Corridor})} with a map-absent chair. 
    After traversing approximately $10\mathrm{\,m}$ in a dynamic environment, the robot encounters a chair placed at the corridor center that is not represented in the pre-collected goal images, and thus must be avoided only by visual observation. This scenario reflects everyday human-space disturbances such as moved furniture, crossing pedestrians, and people stepping away.
    \item \textbf{Long-range navigation (\emph{Long-range})}. 
    The robot follows an approximately $50\mathrm{\,m}$ trajectory that covers about half a circuit of the office, crosses two junctions, and deals with various scene dynamics. 
    The environment contains changes not present in the pre-collected goal images, which evaluates the robustness to appearance shifts and long-horizon navigation.
\end{enumerate}

\begin{table*}[t]
  \centering
  \setlength{\tabcolsep}{4pt}          
  \renewcommand{\arraystretch}{1.4}
  \caption{\textbf{Benchmark using Offline Data.} ADE, FDE, and DTW are the error metrics, where lower values are better.}
  \label{tab:offline-benchmark}
  \resizebox{\textwidth}{!}{%
  \begin{tabular}{lccc|ccc ccc ccc ccc}
    \toprule
    \multirow{3}{*}{Method} &
    \multicolumn{3}{c|}{\textit{F.T. Dataset}} &
    \multicolumn{12}{c}{\textit{NoMaD Dataset}} \\
    \cmidrule(lr){2-4}\cmidrule(lr){5-16}
      & \multicolumn{3}{c|}{RealHSRNav} &
        \multicolumn{3}{c}{Recon~\cite{shah2021rapid}} &
        \multicolumn{3}{c}{GoStanford~\cite{hirose2019deep}} &
        \multicolumn{3}{c}{Sacson~\cite{hirose2023sacson}} &
        \multicolumn{3}{c}{Scand~\cite{karnan2022socially}} \\
    \cmidrule(lr){2-4}
    \cmidrule(lr){5-7}\cmidrule(lr){8-10}\cmidrule(lr){11-13}\cmidrule(lr){14-16}
      & ADE & FDE & DTW
      & ADE & FDE & DTW
      & ADE & FDE & DTW
      & ADE & FDE & DTW
      & ADE & FDE & DTW \\
    \midrule
    \makecell{NoMaD} &
      \runup{1.326} & \runup{2.160} & \runup{0.917} &
      \runup{1.691} & \best{2.996} & \best{1.301} &
      2.267 & \runup{4.448} & \runup{1.888} &
      \runup{2.508} & \best{4.285} & 2.003 &
      2.035 & \runup{2.990} & \runup{1.283} \\
    \makecell{NoMaD-FT} &
      2.138 & 5.484 & 2.255 &
      1.810 & 4.775 & 1.956 &
      \runup{2.097} & 5.436 & 2.216 &
      \best{1.861} & 4.435 & \best{1.913} &
      \best{1.622} & 4.115 & 1.659 \\
    \makecell{NoMaD-EF} &
      1.897 & 4.244 & 1.674 &
      2.222 & 4.676 & 1.991 &
      2.540 & 6.246 & 2.592 &
      2.737 & 5.497 & 2.455 &
      2.428 & 5.063 & 2.045 \\
    \makecell{\textbf{D-CLING} \\ (Ours)} &
      \best{1.298} & \best{1.443} & \best{0.726} &
      \best{1.502} & \runup{3.037} & \runup{1.312} &
      \best{1.812} & \best{4.275} & \best{1.739} &
      2.521 & \runup{4.385} & \runup{1.929} &
      \runup{1.839} & \best{2.401} & \best{1.065} \\
    \bottomrule
  \end{tabular}}%
\end{table*}

\noindent\textbf{Experimental details.}
We conducted the experiments in our office environment on the Toyota HSR, the same platform used to collect the fine-tuning dataset. 
The linear and angular speeds are limited to \(0.45\,\mathrm{m/s}\) and \(1.0\,\mathrm{rad/s}\).
The policy consumes two sources of context: (1) a short visual history of \(T+1\) frames (the current RGB frame and its \(T\) immediate predecessors), each paired with a per-frame depth estimate; and (2) a topological map encoded as an ordered sequence of goal images captured at uniform spatial intervals during the initial setup of the environment.
The model outputs \(H{+}1\) waypoints, including the current step.
We set \(T=3\) and \(H=7\), following NoMaD~\cite{sridhar2024nomad}.

\newcommand{\whatsSF}{
    A feature of the Toyota HSR that stops the robot and logs its event.
}

\noindent\textbf{Metrics.}
For scenarios (i) and (ii), we run 
$10$ trials each and report the success rate (SR). 
A trial is considered successful if the robot reaches the goal without collisions or human intervention.
For scenario (iii), we ran $5$ trials, and recorded the number of detected \textit{safety triggers} from operator interventions.
We report the mean interventions per trial (lower is better), along with the $95\%$ confidence intervals.

\noindent \textbf{Results.}
Table~\ref{tab:real_world_results} reports the real-robot performance across the three scenarios. 
Our proposal, D-CLING, consistently outperforms the baselines. It achieves the highest success rates in (i) and (ii), and requires far fewer interventions in (iii). 
These gains are attributed to the geometry-awareness provided by dense depth adoption while preserving the diverse action patterns of the model, which in turn improves obstacle avoidance and long-horizon goal reachability (Fig.~\ref{fig:real_world_experiment}).

In contrast, zero-shot NoMaD remains problematic, particularly on (ii), even though the model was originally pre-trained on similar \emph{indoor} datasets~\cite{hirose2023sacson,hirose2019deep}.
We conjecture that this is owing to domain shift stemming from the camera geometry and/or scene appearance. 
Furthermore, NoMaD-FT and NoMaD-EF underperform compared to zero-shot NoMaD in (i) and (iii), although they are trained on in-domain data.
In particular, in NoMaD-EF, where the learning for a novel domain is forcibly applied to the RGB-pre-trained policy, the most frequent intervention is required to execute scenario (iii).
We anticipate that input sequences from a novel modality, that is, the dense depth, have eroded pre-trained knowledge.

\noindent \textbf{Limitations.}
We observed the following failure case in our method:
even if the robot successfully avoids the collision \emph{initially}, it immediately returns to the original path and hits the obstacles.
We hypothesize that this failure stems from the limited awareness of the temporal context.
As our NoMaD-based policy conditions on only four (current plus three past) frames without persistent memory mechanisms, obstacles that move off-screen can no longer be tracked by the policy.
This effect became more pronounced in the HSR setup, where a narrower field of view camera is employed than the fisheye camera~\cite{sridhar2024nomad}, further reducing the visible workspace.
Maintaining awareness of the off-screen state via multi-frame fusion or auxiliary sensing may be promising for future studies to reduce such collisions.

\subsection{Offline Evaluations}
To further analyze the impact of our proposal, we evaluate the capability of action prediction, a key subtask of navigation, in an offline setup.
As the action prediction corresponds to the regression of future waypoints in a \textit{metric}-space, a higher awareness of geometry is expected to reflect a more accurate prediction.
For the proposed model, the fine-tuning dataset RealHSRNav and the NoMaD pre-training dataset~\cite{hirose2023sacson,shah2021rapid, hirose2019deep, karnan2022socially} 
are used to verify (1) the prediction capability and (2) the preservation of \emph{previously learned} domain knowledge.
The depth maps for the NoMaD datasets are synthesized using Depth-Anything V2~\cite{yang2024depth} to enable RGB-D methods. 

\noindent \textbf{Experimental details.}
In all experiments, we use a \(T{+}1=4\) frame history (from \(t-3\) to \(t\)) and generate \(H{+}1=8\) waypoints including the current step (that is, \(t,\dots,t{+}7\)), consistent with the real-world experiments.
\noindent
For comprehensiveness, $100$ independent sequences were randomly sampled per dataset and the error score (detailed later) was averaged across the sampled sequences. 
In each trial: (1) a fixed random seed was set for reproducibility; (2) an observation window of $T+1=4$ frames and a goal image were randomly sampled, and a single forward pass was performed to predict the following $H+1=8$ actions; (4) the metrics were logged; and (5) the process proceeded to the next trial with a newly sampled observation window and goal pair.

\noindent\textbf{Metrics.}
We report \emph{Average Displacement Error} (ADE) and \emph{Final Displacement Error} (FDE), where ADE is the mean Euclidean position error over the prediction horizon and FDE is the Euclidean error at the terminal step \(t{+}H\).
To account for temporal misalignments while assessing trajectory fidelity, we also report the (normalized) Dynamic Time Warping (nDTW) distance~\cite{ilharco2019general}, which has been shown to correlate well with human judgment of trajectory similarity.

\begin{figure}[tb]
  \centering

  \begin{minipage}[t]{0.85\columnwidth}
    \centering
    \includegraphics[keepaspectratio,width=\linewidth]{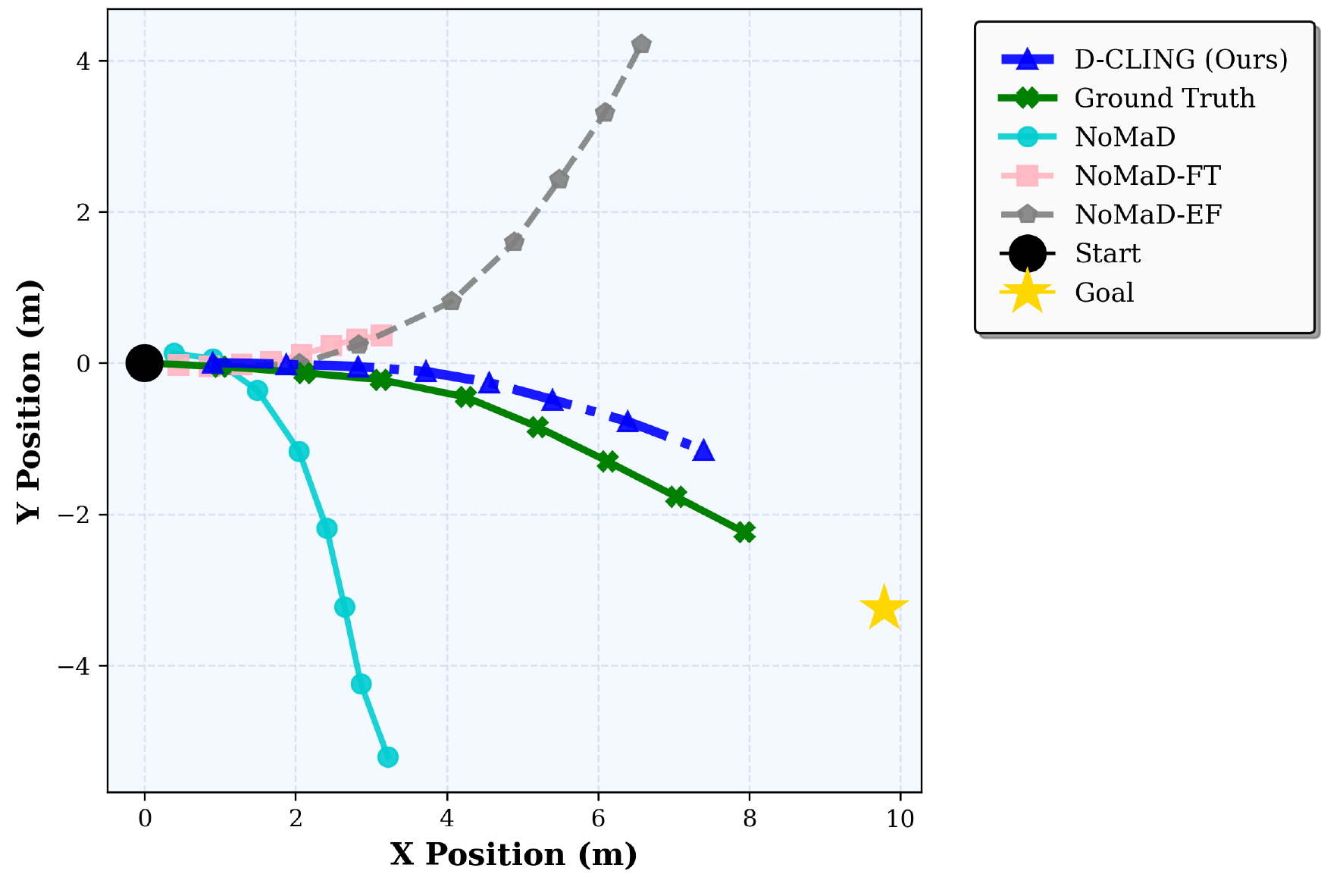}
    \\ {\footnotesize (a) RealNaVHSR.}
    \label{fig:realhsrnav_offline}
  \end{minipage}

  \vspace{0.9em}

  \begin{minipage}[t]{0.95\columnwidth}
    \centering
    \includegraphics[keepaspectratio,width=\linewidth]{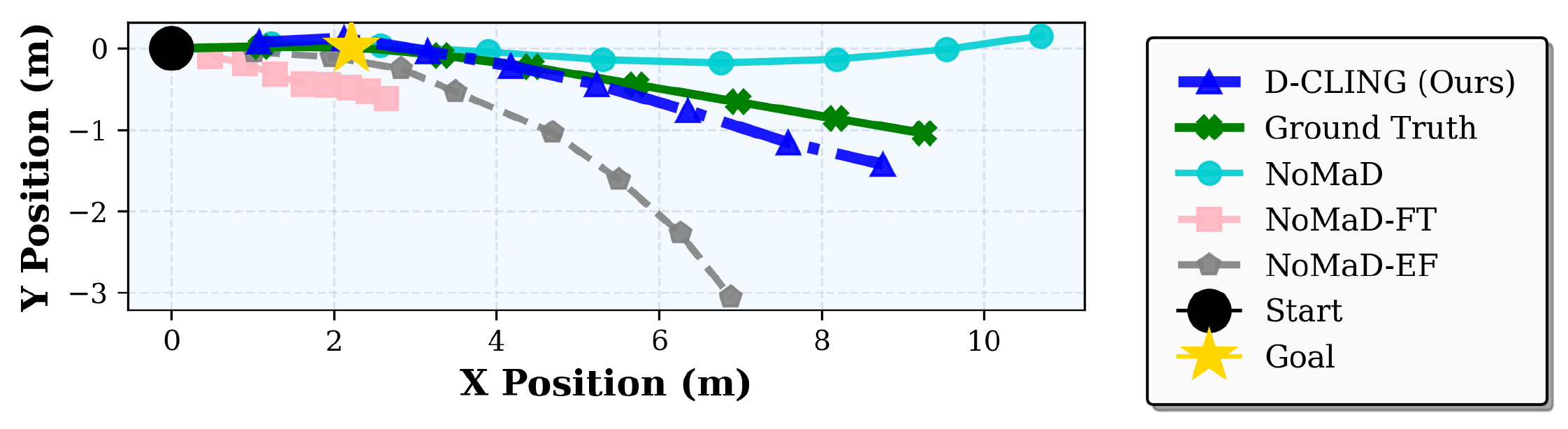}
    \\ {\footnotesize (b) NoMaD Dataset (GoStanford~\cite{hirose2019deep})}
    \label{fig:overview_approach}
    \end{minipage}
  \caption{\textbf{Top-view of the predicted waypoint sequences}. Waypoint predictions of length $H+1=8$ generated from frame windows of length $T+1=4$ on a representative subset of the offline evaluation set. Each panel overlays D-CLING (blue) and the baselines of NoMaD (cyan), NoMaD-FT (pink), and NoMaD-EF (gray) with the ground-truth path in green and start/goal markers. D-CLING most closely follows the ground truth and maintains the most consistent heading toward the goal image point.}
  \label{fig:combined_offline_overview}
\end{figure}

\noindent\textbf{Results and discussion.}
Table~\ref{tab:offline-benchmark} shows the quantitative results, and Figure~\ref{fig:combined_offline_overview} shows the qualitative results of the evaluation. 
On RealHSRNav (the dataset used for fine-tuning), D-CLING attains the lowest ADE, FDE, and DTW, outperforming the other baselines. 
This indicates that our proposed dense depth adaptation strategy offers the best future action prediction in the target environment.
Intriguingly, the evaluation \emph{on NoMaD datasets} demonstrates that fine-tuned D-CLING \emph{on RealHSRNav} provides competitive or \emph{even better} performance compared with the zero-shot NoMaD.
We conjecture that our proposed strategy improved transferability to various scene domains, beyond the fine-tuning domain.

In contrast, NoMaD-FT and NoMaD-EF, which lack a specific mechanism to alleviate catastrophic forgetting, yield inferior results on most domains compared with ours and the zero-shot model.
These findings align with the discussion in Section~\ref{sec:real_world_experiments}, where fine-tuning \emph{rather degraded} the original zero-shot capability.

\subsection{Benefits of Depth Conditioning (RGB vs RGB-D)}

Finally, we investigated the effect of using depth, rather than RGB, given the conditioning strategy that D-CLING employed.
We expected that the comparison between those two conditions would more clearly elucidate the importance of the geometric cue for novel-domain learning. 
Specifically, we implemented the RGB baseline by (1) dropping the depth-input channel from D-CLING, and (2) training with the same protocol (Section ~\ref{sec:depth_conditioned_branch}).

\noindent\textbf{Scenarios.}
We compared the two strategies across two scenarios, 
which are fundamental to real-world navigation:

\begin{enumerate}[label=(\roman*)]
    \item \textbf{Single obstacle avoidance (\emph{Single Obstacle}).}
    The robot traverses a corridor while avoiding a single stationary chair placed along its route.This scenario is similar to the \textit{Dynamic Corridor} in Section~\ref{sec:real_world_experiments} and to the scenarios contained in the training dataset.

    \item \textbf{Multi obstacle avoidance (\emph{Multi Obstacle}).}
    The robot navigates through a $15\,\mathrm{m} \times 5\,\mathrm{m}$ area in which three obstacles are placed at approximately uniform intervals in an alternating left-right arrangement along its direction of travel. The robot must avoid these obstacles while maintaining forward progress.
    Notably, this obstacle configuration is not represented in the training dataset and therefore lies outside the training distribution.
\end{enumerate}

\noindent\textbf{Metrics. }For scenarios (i) and (ii), we run 10 trials each and
report the success rate (SR). 

\noindent\textbf{Results and discussion.}
Table~\ref{tab:benefit_depth_condtioning} reports the real-robot results for the two scenarios. \textbf{D-CLING outperforms the RGB baseline} in both cases. The largest gain is observed in scenario~(ii), which is not represented in the training data.
Contrarily, the baseline in (ii) typically stemmed from a delay in avoiding action. 
These results support the hypothesis that depth conditioning in our proposed method strengthens geometric awareness and enables more robust navigation, even in a novel configuration.

\begin{figure}[t]
  \centering
  \begin{minipage}[t]{0.48\columnwidth}
    \centering
    \includegraphics[width=\linewidth]{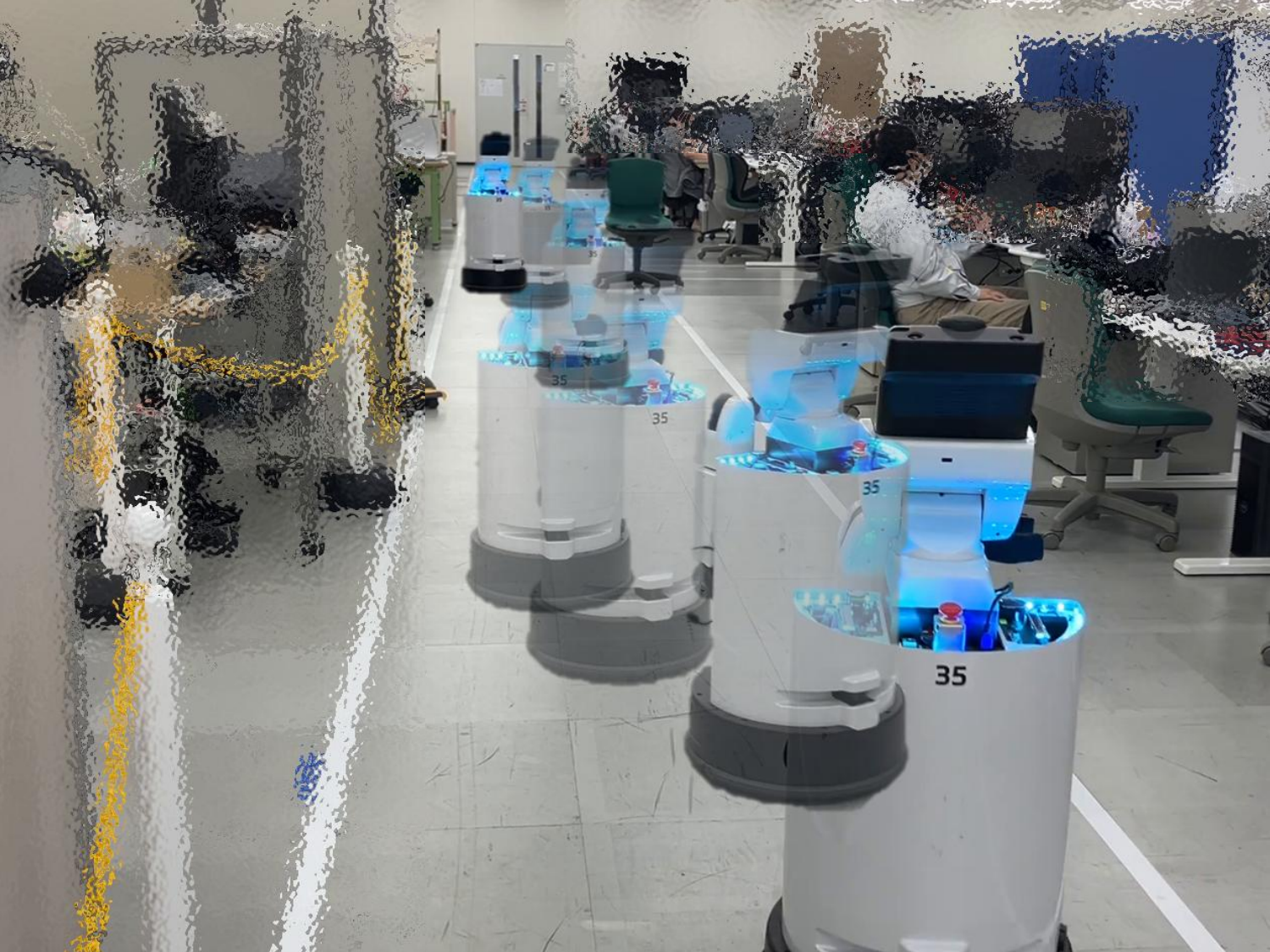}
    
    {\footnotesize (i) \textit{Single Obstacle}}
  \end{minipage}\hfill
  \begin{minipage}[t]{0.48\columnwidth}
    \centering
    \includegraphics[width=\linewidth]{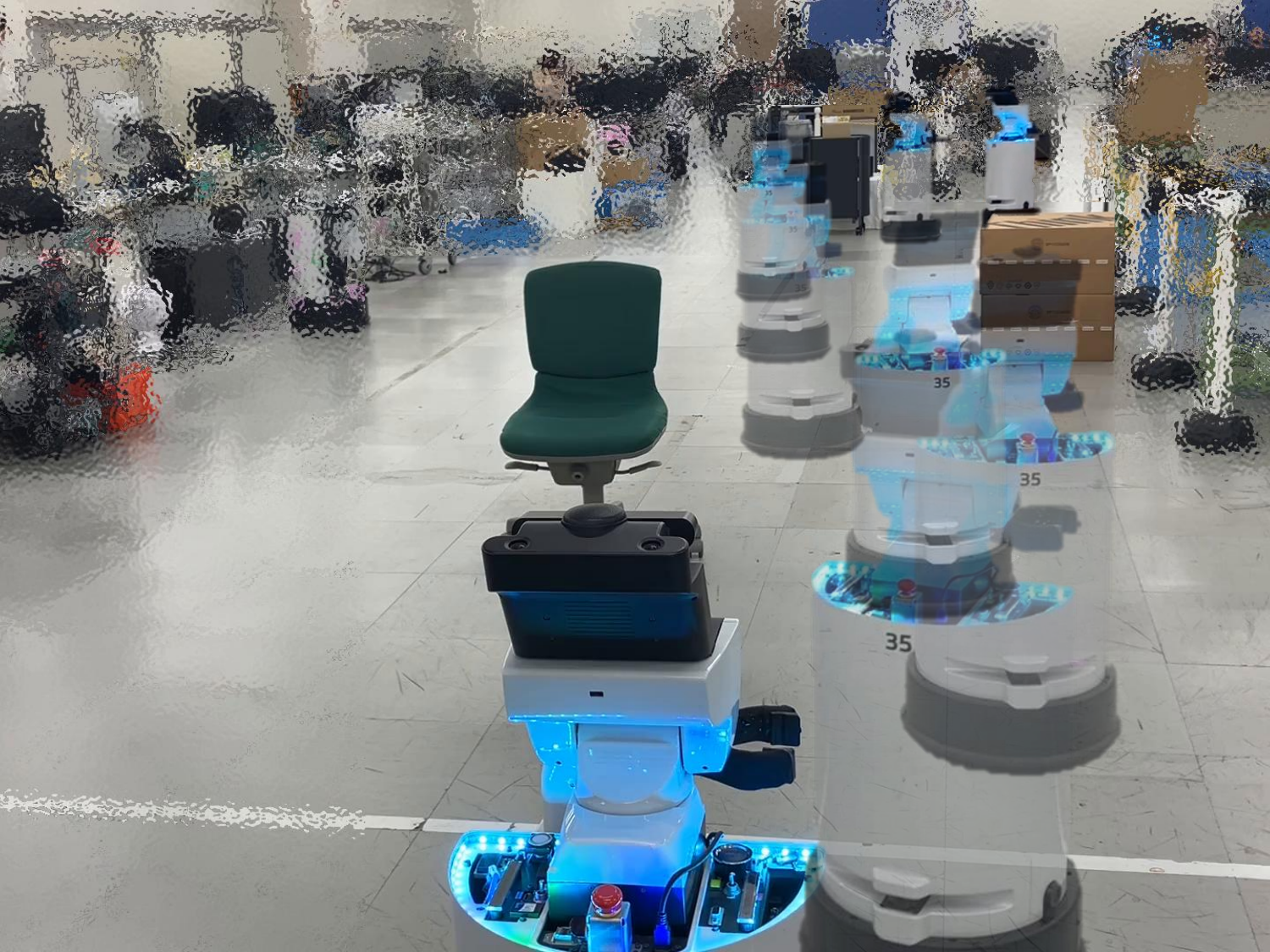}
    
    {\footnotesize (ii) \textit{Multi Obstacle}}
  \end{minipage}
  \caption{\textbf{Representative real-world navigation examples of the RGB-D modality (Ours) in two scenarios.}
(i) \emph{Single Obstacle}: corridor traversal with a single stationary chair.
(ii) \emph{Multi Obstacle}: avoidance of three obstacles in a zig-zag trajectory.}
  \label{fig:depth_vs_rgb_realworld}
\end{figure}



\begin{table}[t]
  \centering
  \normalsize
  \caption{\textbf{RGB vs. RGB-D Conditioning.} Success rate (SR) for two scenarios.}
  \label{tab:benefit_depth_condtioning}

  \resizebox{0.97\linewidth}{!}{
  \begin{tabular}{c c c}
    \toprule
    Modality
    & \shortstack[c]{(i) \textit{Single Obstacle}\\\rule{1.6cm}{0.35pt}\\SR(\%)$\uparrow$}
    & \shortstack[c]{(ii) \textit{Multi Obstacle}\\\rule{1.6cm}{0.35pt}\\SR(\%)$\uparrow$} \\
    \midrule
    RGB & 60 & 10 \\
    \textbf{RGB-D (Ours)} & \textbf{80} & \textbf{100} \\
    \bottomrule
  \end{tabular}
  }
\end{table}

\section{Conclusion and Future Work}
The zero-shot adoption of NFMs suffers from issues such as novel scene observations and camera parameters. 
However, fine-tuning on a limited in-domain dataset is insufficient to overcome these issues.
Furthermore, typical fine-tuning hinders the diverse action generation of pre-trained behavior, which is crucial for various real-world navigation tasks.
This study proposed D-CLING, a prior-preserving and depth-conditioning strategy for NFM fine-tuning to enable novel scenario learning.
The comprehensive analysis demonstrated the efficacy of the proposed method for robust navigation, as well as higher accuracy in action prediction. 
Moreover, the proposed method enables more accurate action prediction beyond the fine-tuned domains, thereby further improving the zero-shot performance of NFMs.

The experiments in this study were limited to NoMaD, which is one example of an NFM. The proposed framework was designed in a largely model-agnostic manner and could potentially be applied to other NFMs. An important direction for future work is the evaluation of the proposed method across a broader range of NFMs.

\section{Acknowledgments}
The authors used ChatGPT for limited language editing and grammar checking, and reviewed all AI-assisted text.

\printbibliography

@String(CVPR= {IEEE Conf. Comput. Vis. Pattern Recog.})

@String(NIPS= {Adv. Neural Inform. Process. Syst.})

@String(ICLR = {Int. Conf. Learn. Represent.})

@String(IJCAI = {IJCAI})

@String(CVPR  = {CVPR})

@String(NIPS  = {NeurIPS})

@String(ICLR  = {ICLR})

@String(AAAI = {AAAI})

@String(IROS = {IROS})

@String(ICRA = {ICRA})

@String(arXiv = {arXiv})

@String(RAL = {IEEE RA-L})

@String(CoRL  = {CoRL})

@String(RSS  = {RSS})

@inproceedings{zhu2017,
  title     = {Target-driven visual navigation in indoor scenes using deep reinforcement learning},
  author    = {Zhu, Yuke and Mottaghi, Roozbeh and Kolve, Eric and Lim, Joseph J and Gupta, Abhinav and Fei-Fei, Li and Farhadi, Ali},
  booktitle = ICRA,
  memo      = {
               booktitle    = {2017 IEEE international conference on robotics and automation (ICRA)},
               organization = {IEEE}
               },
  pages     = {3357--3364},
  year      = {2017}
}

@article{wijmans2019dd,
  title   = {Dd-ppo: Learning near-perfect pointgoal navigators from 2.5 billion frames},
  author  = {Wijmans, Erik and Kadian, Abhishek and Morcos, Ari and Lee, Stefan and Essa, Irfan and Parikh, Devi and Savva, Manolis and Batra, Dhruv},
  journal = arXiv,
  memo    = {
             journal = {arXiv preprint arXiv:1911.00357},
             },
  year    = {2019}
}

@inproceedings{anderson2018vision,
  title     = {Vision-and-language navigation: Interpreting visually-grounded navigation instructions in real environments},
  author    = {Anderson, Peter and Wu, Qi and Teney, Damien and Bruce, Jake and Johnson, Mark and S{\"u}nderhauf, Niko and Reid, Ian and Gould, Stephen and Van Den Hengel, Anton},
  booktitle = CVPR,
  memo      = {
               booktitle = {Proceedings of the IEEE conference on computer vision and pattern recognition},
               },
  pages     = {3674--3683},
  year      = {2018}
}

@inproceedings{o2024open,
  title={Open x-embodiment: Robotic learning datasets and rt-x models: Open x-embodiment collaboration 0},
  author={O’Neill, Abby and Rehman, Abdul and Maddukuri, Abhiram and Gupta, Abhishek and Padalkar, Abhishek and Lee, Abraham and Pooley, Acorn and Gupta, Agrim and Mandlekar, Ajay and Jain, Ajinkya and others},
  booktitle=ICRA,
  pages={6892--6903},
  year={2024},
  organization={IEEE}
}

@article{intelligence2025pi_,
  title   = {$\pi_{0.5}$: a Vision-Language-Action Model with Open-World Generalization},
  author  = {Black, Kevin and Brown, Noah and Darpinian, James and Dhabalia, Karan and Driess, Danny and Esmail, Adnan and Equi, Michael and Finn, Chelsea and Fusai, Niccolo and others},
  journal = arXiv,
  memo    = {
             author  = {Intelligence, Physical and Black, Kevin and Brown, Noah and Darpinian, James and Dhabalia, Karan and Driess, Danny and Esmail, Adnan and Equi, Michael and Finn, Chelsea and Fusai, Niccolo and others},
             journal = {arXiv preprint arXiv:2504.16054},
             },
  year    = {2025}
}

@article{bjorck2025gr00t,
  title   = {Gr00t n1: An open foundation model for generalist humanoid robots},
  author  = {Bjorck, Johan and Casta{\~n}eda, Fernando and Cherniadev, Nikita and Da, Xingye and Ding, Runyu and Fan, Linxi and Fang, Yu and Fox, Dieter and Hu, Fengyuan and Huang, Spencer and others},
  journal = arXiv,
  memo    = {
             journal = {arXiv preprint arXiv:2503.14734},
             },
  year    = {2025}
}

@article{shah2023vint,
  title   = {ViNT: A foundation model for visual navigation},
  author  = {Shah, Dhruv and Sridhar, Ajay and Dashora, Nitish and Stachowicz, Kyle and Black, Kevin and Hirose, Noriaki and Levine, Sergey},
  journal = arXiv,
  memo    = {
             journal={arXiv preprint arXiv:2306.14846},
             },
  year    = {2023}
}

@inproceedings{sridhar2024nomad,
  title     = {Nomad: Goal masked diffusion policies for navigation and exploration},
  author    = {Sridhar, Ajay and Shah, Dhruv and Glossop, Catherine and Levine, Sergey},
  booktitle = ICRA,
  pages     = {63--70},
  year      = {2024},
  memo      = {
               organization = {IEEE},
               booktitle    = {2024 IEEE International Conference on Robotics and Automation (ICRA)},
               }
}

@article{wan2025pig,
  title   = {PIG-Nav: Key Insights for Pretrained Image Goal Navigation Models},
  author  = {Wan, Jiansong and Zhou, Chengming and Liu, Jinkua and Huang, Xiangge and Chen, Xiaoyu and Yi, Xiaohan and Yang, Qisen and Zhu, Baiting and Cai, Xin-Qiang and Liu, Lixing and others},
  journal = arXiv,
  memo    = {
             journal = {arXiv preprint arXiv:2507.17220},
             },
  year    = {2025}
}

@inproceedings{stachowicz2024lifelong,
  title     = {Lifelong autonomous improvement of navigation foundation models in the wild},
  author    = {Stachowicz, Kyle and Ignatova, Lydia and Levine, Sergey},
  booktitle = CoRL,
  memo      = {
               booktitle = {8th Annual Conference on Robot Learning},
               },
  year      = {2024}
}

@article{kim2025enhancing,
  title   = {Enhancing Safety of Foundation Models for Visual Navigation through Collision Avoidance via Repulsive Estimation},
  author  = {Kim, Joonkyung and Sim, Joonyeol and Kim, Woojun and Sycara, Katia and Nam, Changjoo},
  journal = arXiv,
  memo    = {
             journal = {arXiv preprint arXiv:2506.03834},
             },
  year    = {2025}
}

@inproceedings{zhang2023adding,
  title     = {Adding conditional control to text-to-image diffusion models},
  author    = {Zhang, Lvmin and Rao, Anyi and Agrawala, Maneesh},
  booktitle = CVPR,
  pages     = {3836--3847},
  year      = {2023},
  memo      = {
               booktitle = {Proceedings of the IEEE/CVF international conference on computer vision},
               }
}

@article{Yamamoto2019,
  author    = {Yamamoto, Takashi and Terada, Koji and Ochiai, Akiyoshi and Saito, Fuminori and Asahara, Yoshiaki and Murase, Kazuto},
  doi       = {10.1186/s40648-019-0132-3},
  issn      = {2197-4225},
  journal   = {ROBOMECH J.},
  memo      = {
               journal   = {ROBOMECH Journal},
               },
  month     = {dec},
  number    = {1},
  pages     = {4},
  publisher = {Springer International Publishing},
  title     = {{Development of Human Support Robot as the research platform of a domestic mobile manipulator}},
  volume    = {6},
  year      = {2019}
}

@article{shankar2022learned,
  title   = {A learned stereo depth system for robotic manipulation in homes},
  author  = {Shankar, Krishna and Tjersland, Mark and Ma, Jeremy and Stone, Kevin and Bajracharya, Max},
  journal = RAL,
  memo    = {
             journal   = {IEEE Robotics and Automation Letters},
             publisher = {IEEE},
             },
  volume  = {7},
  number  = {2},
  pages   = {2305--2312},
  year    = {2022}
}

@article{Gasparino2022,
  author  = {Gasparino, Mateus V and Sivakumar, Arun N and Liu, Yixiao and Velasquez, Andres E B and Higuti, Vitor A H and Rogers, John and Tran, Huy and Chowdhary, Girish},
  doi     = {10.1109/LRA.2022.3193464},
  issn    = {2377-3766},
  journal = RAL,
  number  = {4},
  pages   = {10651--10658},
  title   = {{WayFAST: Navigation With Predictive Traversability in the Field}},
  url     = {https://ieeexplore.ieee.org/document/9839522/},
  volume  = {7},
  year    = {2022},
  memo    = {
             journal   = {IEEE Robotics and Automation Letters},
             publisher = {IEEE},
             }
}

@article{bousmalis2023robocat,
  title   = {Robocat: A self-improving generalist agent for robotic manipulation},
  author  = {Bousmalis, Konstantinos and Vezzani, Giulia and Rao, Dushyant and Devin, Coline and Lee, Alex X and Bauz{\'a}, Maria and Davchev, Todor and Zhou, Yuxiang and Gupta, Agrim and Raju, Akhil and others},
  journal = arXiv,
  memo    = {
             journal = {arXiv preprint arXiv:2306.11706},
             },
  year    = {2023}
}

@inproceedings{ramrakhya2022habitat,
  title     = {Habitat-web: Learning embodied object-search strategies from human demonstrations at scale},
  author    = {Ramrakhya, Ram and Undersander, Eric and Batra, Dhruv and Das, Abhishek},
  booktitle = CVPR,
  memo      = {
               booktitle = {Proceedings of the IEEE/CVF conference on computer vision and pattern recognition},
               },
  pages     = {5173--5183},
  year      = {2022}
}

@inproceedings{tai2018socially,
  title     = {Socially compliant navigation through raw depth inputs with generative adversarial imitation learning},
  author    = {Tai, Lei and Zhang, Jingwei and Liu, Ming and Burgard, Wolfram},
  booktitle = ICRA,
  memo      = {
               booktitle    = {2018 IEEE international conference on robotics and automation (ICRA)},
               organization = {IEEE},
               },
  pages     = {1111--1117},
  year      = {2018}
}

@article{gode2024flownav,
  title   = {Flownav: Combining flow matching and depth priors for efficient navigation},
  author  = {Gode, Samiran and Nayak, Abhijeet and Oliveira, D{\'e}bora NP and Krawez, Michael and Schmid, Cordelia and Burgard, Wolfram},
  journal = arXiv,
  memo    = {
             journal = {arXiv preprint arXiv:2411.09524},
             },
  year    = {2024}
}

@article{feng2025image,
  title   = {Image-Goal Navigation Using Refined Feature Guidance and Scene Graph Enhancement},
  author  = {Feng, Zhicheng and Chen, Xieyuanli and Shi, Chenghao and Luo, Lun and Chen, Zhichao and Liu, Yun-Hui and Lu, Huimin},
  journal = arXiv,
  memo    = {
             journal = {arXiv preprint arXiv:2503.10986},
             },
  year    = {2025}
}

@inproceedings{liu2025citywalker,
  title     = {Citywalker: Learning embodied urban navigation from web-scale videos},
  author    = {Liu, Xinhao and Li, Jintong and Jiang, Yicheng and Sujay, Niranjan and Yang, Zhicheng and Zhang, Juexiao and Abanes, John and Zhang, Jing and Feng, Chen},
  booktitle = CVPR,
  memo      = {
               booktitle = {Proceedings of the Computer Vision and Pattern Recognition Conference},
               },
  pages     = {6875--6885},
  year      = {2025}
}

@inproceedings{thoma2019mapping,
  title     = {Mapping, localization and path planning for image-based navigation using visual features and map},
  author    = {Thoma, Janine and Paudel, Danda Pani and Chhatkuli, Ajad and Probst, Thomas and Gool, Luc Van},
  booktitle = CVPR,
  memo      = {
               booktitle = {Proceedings of the Computer Vision and Pattern Recognition Conference},
               },
  pages     = {7383--7391},
  year      = {2019}
}

@inproceedings{kamath2023new,
  title     = {A new path: Scaling vision-and-language navigation with synthetic instructions and imitation learning},
  author    = {Kamath, Aishwarya and Anderson, Peter and Wang, Su and Koh, Jing Yu and Ku, Alexander and Waters, Austin and Yang, Yinfei and Baldridge, Jason and Parekh, Zarana},
  booktitle = CVPR,
  memo      = {
               booktitle = {Proceedings of the IEEE/CVF conference on computer vision and pattern recognition},
               },
  pages     = {10813--10823},
  year      = {2023}
}

@article{song2024vlm,
  title     = {Vlm-social-nav: Socially aware robot navigation through scoring using vision-language models},
  author    = {Song, Daeun and Liang, Jing and Payandeh, Amirreza and Raj, Amir Hossain and Xiao, Xuesu and Manocha, Dinesh},
  journal   = RAL,
  memo      = {
               journal   = {IEEE Robotics and Automation Letters},
               },
  year      = {2024},
  publisher = {IEEE}
}

@article{hirose2024lelan,
  title   = {Lelan: Learning a language-conditioned navigation policy from in-the-wild videos},
  author  = {Hirose, Noriaki and Glossop, Catherine and Sridhar, Ajay and Shah, Dhruv and Mees, Oier and Levine, Sergey},
  journal = arXiv,
  memo    = {
             journal = {arXiv preprint arXiv:2410.03603},
             },
  year    = {2024}
}

@article{hirose2023sacson,
  title   = {Sacson: Scalable autonomous control for social navigation},
  author  = {Hirose, Noriaki and Shah, Dhruv and Sridhar, Ajay and Levine, Sergey},
  journal = RAL,
  volume  = {9},
  memo    = {
             journal   = {IEEE Robotics and Automation Letters},
             publisher = {IEEE},
             },
  number  = {1},
  pages   = {49--56},
  year    = {2023}
}

@article{cai2025navdp,
  title   = {NavDP: Learning Sim-to-Real Navigation Diffusion Policy with Privileged Information Guidance},
  author  = {Cai, Wenzhe and Peng, Jiaqi and Yang, Yuqiang and Zhang, Yujian and Wei, Meng and Wang, Hanqing and Chen, Yilun and Wang, Tai and Pang, Jiangmiao},
  journal = arXiv,
  memo    = {
             journal = {arXiv preprint arXiv:2505.08712},
             },
  year    = {2025}
}

@article{wang2025x,
  title   = {X-Nav: Learning End-to-End Cross-Embodiment Navigation for Mobile Robots},
  author  = {Wang, Haitong and Tan, Aaron Hao and Fung, Angus and Nejat, Goldie},
  journal = arXiv,
  memo    = {
             journal = {arXiv preprint arXiv:2507.14731},
             },
  year    = {2025}
}

@article{shah2021rapid,
  title   = {Rapid exploration for open-world navigation with latent goal models},
  author  = {Shah, Dhruv and Eysenbach, Benjamin and Kahn, Gregory and Rhinehart, Nicholas and Levine, Sergey},
  journal = arXiv,
  memo    = {
             journal = {arXiv preprint arXiv:2104.05859},
             },
  year    = {2021}
}

@article{karnan2022socially,
  title     = {Socially compliant navigation dataset (scand): A large-scale dataset of demonstrations for social navigation},
  author    = {Karnan, Haresh and Nair, Anirudh and Xiao, Xuesu and Warnell, Garrett and Pirk, S{\"o}ren and Toshev, Alexander and Hart, Justin and Biswas, Joydeep and Stone, Peter},
  journal   = RAL,
  memo      = {
               journal   = {IEEE Robotics and Automation Letters},
               },
  volume    = {7},
  number    = {4},
  pages     = {11807--11814},
  year      = {2022},
  publisher = {IEEE}
}

@article{hirose2019deep,
  title     = {Deep visual mpc-policy learning for navigation},
  author    = {Hirose, Noriaki and Xia, Fei and Mart{\'\i}n-Mart{\'\i}n, Roberto and Sadeghian, Amir and Savarese, Silvio},
  journal   = RAL,
  memo      = {
               journal   = {IEEE Robotics and Automation Letters},
               },
  volume    = {4},
  number    = {4},
  pages     = {3184--3191},
  year      = {2019},
  publisher = {IEEE}
}

@article{yang2024depth,
  title   = {Depth anything v2},
  author  = {Yang, Lihe and Kang, Bingyi and Huang, Zilong and Zhao, Zhen and Xu, Xiaogang and Feng, Jiashi and Zhao, Hengshuang},
  journal = NIPS,
  memo    = {
             journal = {Advances in Neural Information Processing Systems},
             },
  volume  = {37},
  pages   = {21875--21911},
  year    = {2024}
}

@article{ilharco2019general,
  title   = {General evaluation for instruction conditioned navigation using dynamic time warping},
  author  = {Ilharco, Gabriel and Jain, Vihan and Ku, Alexander and Ie, Eugene and Baldridge, Jason},
  journal = arXiv,
  memo    = {
             journal = {arXiv preprint arXiv:1907.05446},
             },
  year    = {2019}
}

@article{loshchilov2017decoupled,
  title={Decoupled weight decay regularization},
  author={Loshchilov, Ilya and Hutter, Frank},
  journal=arXiv,
   memo={arXiv preprint arXiv:1711.05101},
  year={2017}
}

@inproceedings{zhang2025creste,
      title={CREStE: Scalable Mapless Navigation with Internet Scale Priors and Counterfactual Guidance}, 
      author={Arthur Zhang and Harshit Sikchi and Amy Zhang and Joydeep Biswas},
      booktitle=RSS,
      year={2025},
      memo = {
      booktitle={Robotics: Science and Systems (RSS)},
      }
}

@article{barreiros2025careful,
  title={A careful examination of large behavior models for multitask dexterous manipulation},
  author={TRI LBM Team},
    journal = arXiv,
    memo    = {
             journal = {arXiv preprint arXiv:2507.05331},
             },
  year={2025}
}

@inproceedings{lu2024learning,
  title={Learning generalizable manipulation policy with adapter-based parameter fine-tuning},
  author={Lu, Kai and Ly, Kim Tien and Hebberd, William and Zhou, Kaichen and Havoutis, Ioannis and Markham, Andrew},
  booktitle = IROS,
  memo      = {
               booktitle    = {024 IEEE/RSJ International Conference on Intelligent Robots and Systems (IROS)},
               organization = {IEEE},
            
               },
  year={2024},
page={13510--13517},
}

@inproceedings{Liu2024,
 author = {Zuxin Liu and Jesse Zhang and Kavosh Asadi and Yao Liu and Ding Zhao and Shoham Sabach and Rasool Fakoor},
 title = {TAIL: Task-specific adapters for imitation learning with large pretrained models},
 booktitle = ICLR,
 year = {2024},
  memo ={url = {https://www.amazon.science/publications/tail-task-specific-adapters-for-imitation-learning-with-large-pretrained-models}},
}

@article{sharmalossless,
  title={Lossless Adaptation of Pretrained Vision Models For Robotic Manipulation},
  author={Sharma, Mohit and Fantacci, Claudio and Zhou, Yuxiang and Koppula, Skanda and Heess, Nicolas and Scholz, Jon and Aytar, Yusuf},
     journal= arXiv,
    memo    = {arXiv preprint arXiv:2304.06600},
  year={2023}
}

@article{mur2015orb,
  title={ORB-SLAM: A versatile and accurate monocular SLAM system},
  author={Mur-Artal, Raul and Montiel, Jose Maria Martinez and Tardos, Juan D},
  journal={IEEE transactions on robotics},
  volume={31},
  number={5},
  pages={1147--1163},
  year={2015},
  publisher={IEEE}
}

@article{honda2025gsplatvnm,
  title={GSplatVNM: Point-of-View Synthesis for Visual Navigation Models Using Gaussian Splatting},
  author={Honda, Kohei and Ishita, Takeshi and Yoshimura, Yasuhiro and Yonetani, Ryo},
  journal=arXiv,
 memo ={arXiv preprint arXiv:2503.05152},
  year={2025}
}

@inproceedings{ijcai2020p124,
  title     = {Diagnosing the Environment Bias in Vision-and-Language Navigation},
  author    = {Zhang, Yubo and Tan, Hao and Bansal, Mohit},
  booktitle = IJCAI,
 memo ={
  publisher = {International Joint Conferences on Artificial Intelligence Organization},
  editor    = {Christian Bessiere},
   month     = {7},
  note      = {Main track},
  doi       = {10.24963/ijcai.2020/124},
  url       = {https://doi.org/10.24963/ijcai.2020/124}, 
     },
  pages     = {890--897},
year      = {2020},

}

@inproceedings{tang2022mono,
  title={Monocular Camera-Based Point-Goal Navigation by Learning Depth Channel and Cross-Modality Pyramid Fusion},
  author={Tianqi Tang, Heming Du, Xin Yu and Yi Yang},
  booktitle={AAAI},
  pages={36(5), 5422-5430},
  year={2022}
}

@inproceedings{chaplot2020learning,
  title={Learning To Explore Using Active Neural SLAM},
  author={Chaplot, Devendra Singh and Gandhi, Dhiraj and Gupta,
          Saurabh and Gupta, Abhinav and Salakhutdinov, Ruslan},
  booktitle=ICLR,
  year={2020}}

@inproceedings{qiu2020learning,
  author={Y. {Qiu} and A. {Pal} and H. I. {Christensen}},
  booktitle=CORL, 
  title={Learning hierarchical relationships for object-goal navigation}, 
  year={2020}}

@article{suomela2026data,
  title={Data Scaling for Navigation in Unknown Environments},
  author={Suomela, Lauri and Takahata, Naoki and Arachchige, Sasanka Kuruppu and Edelman, Harry and K{\"a}m{\"a}r{\"a}inen, Joni-Kristian},
  journal=arXiv,
  year={2026}
}

\end{document}